\documentclass[journal]{IEEEtran}
\usepackage[top=0.75in, bottom=1in, left=0.5in, right=0.5in]{geometry}
\usepackage{amsmath,amsfonts,amscd,amssymb}
\usepackage{graphicx}
\usepackage{epstopdf}
\usepackage{overpic}
\usepackage{cancel}
\usepackage{rotating}
\usepackage{url}
\usepackage{caption}
\usepackage{color}
\usepackage{rotating}
\usepackage{multirow}
\usepackage{wrapfig}
\usepackage{mathtools}
\usepackage{subeqnarray}
\usepackage{setspace}
\usepackage{hhline}
\usepackage{palatino} % needs 10
\usepackage[numbers,sort&compress]{natbib}
\usepackage{arydshln}
\usepackage{caption}
\usepackage{subcaption}
\usepackage{float}

\usepackage[bottom,flushmargin,hang,multiple]{footmisc}
\usepackage{lipsum}
\newcommand\blfootnote[1]{%
  \begingroup
  \renewcommand\thefootnote{}\footnote{#1}%
  \addtocounter{footnote}{-1}%
  \endgroup
}

\definecolor{header1}{cmyk}{0,0,0,1}

\DeclareMathOperator*{\argmin}{arg\rm{}min}

\newcommand{\Av}{\mathbf{A}}
\newcommand{\Bv}{\mathbf{B}}
\newcommand{\Wv}{\mathbf{W}}
\newcommand{\Xv}{\mathbf{X}}
\newcommand{\Yv}{\mathbf{Y}}

\newcommand{\Fv}{\mathbf{F}}
\newcommand{\Sigv}{\mathbf{\Sigma}}
\newcommand{\Lamv}{\mathbf{\Lambda}}
\newcommand{\Uv}{\mathbf{U}}
\newcommand{\Vv}{\mathbf{V}}
\newcommand{\Kv}{\mathbf{K}}

\newcommand{\pv}{\mathbf{p}}

\newcommand{\xv}{\mathbf{x}}
\newcommand{\yv}{\mathbf{y}}
\newcommand{\vv}{\mathbf{v}}
\newcommand{\uv}{\mathbf{u}}

\newcommand{\sv}{\mathbf{s}}
\newcommand{\Sv}{\mathbf{S}}
\newcommand{\kv}{\mathbf{k}}

\newcommand{\thetav}{\boldsymbol{\theta}}

\renewcommand{\tilde}{\widetilde}

\newcommand{\diag}{\textbf{diag}}
\newcommand{\T}{^\textsf{T}}

\usepackage{xcolor}
\title{\huge{\textbf{Swarm Modelling with Dynamic Mode Decomposition}}}

\author{\normalsize{Emma Hansen$^{1*}$, Steven L. Brunton$^2$ and Zhuoyuan Song$^3$}\\
\footnotesize{$^1$ Mechanical Engineering, University of Washington, Seattle, WA 98195, United States}\\
\footnotesize{$^2$ Mechanical Engineering, University of Washington, Seattle, WA 98195, United States}\\
\footnotesize{$^3$ Mechanical Engineering, University of Hawai`i at M\={a}noa, Honolulu, HI 96822, United States}
}
\date{}
\begin{document}
\maketitle

\blfootnote{$^*$ Corresponding author (ehansen@math.ubc.edu).}
%%%%%%%%%%%%
%%% ABSTRACT
%%%%%%%%%%%%
\begin{abstract}
Modelling biological or engineering swarms is challenging due to the inherently high dimension of the system, despite the often low-dimensional emergent dynamics. 
Most existing swarm modelling approaches are based on first principles and often result in swarm-specific parameterizations that do not generalize to a broad range of applications. 
In this work, we apply a purely data-driven method to (1) learn local interactions of homogeneous swarms through observation data and to (2) generate similar swarming behaviour using the learned model. 
In particular, a modified version of dynamic mode decomposition with control, called swarmDMD, is developed and tested on the canonical Vicsek swarm model.  
The goal is to use swarmDMD to learn inter-agent interactions that give rise to the observed swarm behaviour.
We show that swarmDMD can faithfully reconstruct the swarm dynamics, and the model learned by swarmDMD provides a short prediction window for data extrapolation with a trade-off between prediction accuracy and prediction horizon.
We also provide a comprehensive analysis on the efficacy of different observation data types on the modelling, where we find that inter-agent distance yields the most accurate models.
We believe the proposed swarmDMD approach will be useful for studying multi-agent systems found in biology, physics, and engineering.\\

\noindent\emph{Keywords--}
Swarms, multi-agent systems, reduced-order models, dynamic mode decomposition, control, optimization
\end{abstract}

\section{Introduction}

Emergent behaviours, such as swarming and flocking, are ubiquitous in natural and engineered systems. Fascinating phenomena including birds flocking, fish schooling, and ants colonizing have intrigued generations of biologists to discover and understand the rules of life~\cite{Bonabeau:99a,Lukeman2010,Bialek2014}. Fueled by the interplay between the intra-swarm and swarm-environment interactions, swarms often exhibit complex multi-scale dynamics, making them an interesting research subject also for physicists and mathematicians. Inspired largely by the biology community, engineers and roboticists study swarms in order to understand the emergence of macroscopic organized behaviours for dynamics modeling and prediction~\cite{Berman2011,Vicsek2012,Brambilla2013,Kelley2013}, to build robotic platforms to directly interact with biological swarms~\cite{Strombom2014,Paranjape2018,Romano2019}, and ultimately to design scalable control and planning algorithms for artificial swarms~\cite{Prorok2011,Rubenstein2014,Chung2018,McGuire2019,Schranz2020,Berlinger2021,Dorigo2021}.

One long-standing challenge in the study of swarms and emergent behaviours is the discovery of optimal interaction laws that lead to various macroscopic phenomena resembling our observations in nature. Although several mathematically simple swarm interaction laws have been proposed to generate compelling swarm or flocking behaviour~\cite{reynolds_flocks_1987,Spears2004,Olfati-saber2006,Hsieh2008,Pimenta2013,Song2017}, there remains a need for a principled approach that allows the extraction of the fundamental interaction law from observations alone. To this end, in other fields data-driven methods have proven effective in identifying unknown dynamics from observation~\cite{schmidt_distilling_2009,brunton_discovering_2016,kutz_dynamic_2016}. However, data-driven discovery of swarm interactions is still nascent~\cite{Huttenrauch:19,Zhong2020}.

Collective behaviour in biological swarms has been studied in depth across organisms ranging from cellular systems \cite{davidson_hierarchical_2021} and midges \cite{Kelley2013}, to jackdaw flocks \cite{ling_collective_2019} and humans playing baseball \cite{fujii_physically-interpretable_2020}. Studies on the decision making processes of biological swarms have shown geometry often plays a roll, resulting in a binary process used across biological scales \cite{sridhar_geometry_2021}; whereas other studies have shown further structure is often needed to make general statements across scales \cite{davidson_hierarchical_2021}. Much research has been conducted on the initiation of large collective movements \cite{ling_collective_2019}, showing that environmental cues play an important role in their emergence \cite{van_der_vaart_environmental_2020}; and on disruptions to collective behaviour showing that genetic mutations in zebra fish can cause reduced cohesion in formations \cite{tang_genetic_2020}. Such studies inform the development and evaluation of methods for determining governing laws of swarm interactions \cite{Sumpter:10}.

Many methods to learn swarm laws take a model-based approach \cite{Sumpter:10}. Some of the simplest models, aimed only at recreating the basic structure of emergent patterns, include the alignment-based Vicsek model and its modifications \cite{vicsek_novel_1995,costanzo_spontaneous_2018}, a stochastic model by Aoki \cite{aoki_simulation_1982}, and attraction-repulsion models \cite{reynolds_flocks_1987,helbing_social_1995}. Although biologically inspired, these models were developed with the aim of recreating similar dynamics to those seen in biological swarms, and not at determining the actual underlying governing equations. Model-based methods that do aim to discover governing equations, often make simplifying assumptions \cite{schaeffer_extracting_2018, sinhuber_equation_2021}. In recent years, machine learning \cite{bhaskar_analyzing_2019} and game theoretic \cite{mei_dynamic_2018} approaches are becoming more widely used.

Fully capturing the governing laws of a swarm system with a model-based approach would likely require in a very complicated model.
This promotes the use of model-free, or data-driven methods, where the hope is that these approaches can accurately capture dynamics without the need for restrictive assumptions. Many model-free approaches exploit known properties of swarms, such as the interchangeability of agents \cite{Huttenrauch:19}, the inherent network dynamical interactions \cite{fujii_physically-interpretable_2020}, and how pairwise distances between agents play an important role in interactions \cite{lu_nonparametric_2019}. 

In our proposed method, we take a fully data-driven approach inspired by dynamic mode decomposition (DMD) \cite{schmid_dynamic_2010,Rowley2009jfm,kutz_dynamic_2016,schmid2011applications,schmid2022dynamic,Brunton2021koopman}, a method originating in fluid dynamics and capable of capturing system dynamics across timescales. Using DMD with control as a starting point \cite{proctor_dynamic_2016}, we exploit the general idea of how swarm agents update their states --- using information about their own state and their neighbours' --- to develop dynamic mode decomposition for swarms (swarmDMD). The resulting modelling framework produces an optimal linear approximation, distinguishing our approach from others which may seek nonlinear estimates. As swarmDMD is a purely data-driven approach, with no assumptions made for specific swarm systems, it is generalisable to many kinds of multi-agent systems, and can use different types of state data that are relevant to the dynamics. For data generated by the Vicsek model, it was shown via a comparison of data types that swarmDMD performs the best with relative position information (i.e., inter-agent distance), resulting in accurate recreations of swarm dynamics during the training period and a window of accurate prediction post-training. 

% individual agent dynamics: \cite{alba_exploring_2020}

% multi-agent (to study bio swarms): \cite{kelley_emergent_2013}, \cite{ling_collective_2019}, \cite{ouellette_goals_2021}, \cite{sridhar_geometry_2021}, \cite{tang_genetic_2020}, \cite{van_der_vaart_environmental_2020}

% multi-agent (model-based, to reproduce dynamics): \cite{bhaskar_analyzing_2019}, \cite{chen_searching_2019}, \cite{davidson_hierarchical_2021}, \cite{mei_dynamic_2018}, \cite{mohagheghi_stable_2020}, \cite{sinhuber_equation_2021}, 

% multi-agent (model-free, to reproduce dynamics): \cite{fujii_physically-interpretable_2020}, \cite{huttenrauch_deep_2019}, \cite{lu_nonparametric_2019}, \cite{schaeffer_extracting_2018} (but also does uses models sometimes), 

% methods: \cite{bradde_pca_2017}, \cite{liang_parameter_2008}, \cite{little_multiscale_2017}, \cite{topaz_topological_2015}, 

The remainder of this paper is organized as follows. Section \ref{sec:prelim} provides an overview of DMD for data-driven approximation of nonlinear dynamics. Section \ref{sec:swarmDMD} introduces swarmDMD and how it is developed from DMD. Section \ref{sec:setup} covers the numerical simulation set-up, including the Vicsek model and modification used for the ground truth swarm data sets, and evaluation metrics for the swarmDMD results. Section \ref{sec:results} discusses the experimental results and major findings.

\section{Preliminaries} \label{sec:prelim}

In this section, we introduce important background material and concepts used in the development of swarmDMD. Beginning with DMD, we then describe the DMD with control variant, on which swarmDMD is based.

We focus on swarm behaviours that are induced by local interactions between swarm agents. These emergent behaviours often give rise to coherent swarm structures that can  persist with inherently low-dimensional dynamics. For such systems, a method like DMD, which captures low-order dynamics in high-order systems, is promising in modelling the dominant modes of the systems and providing a short window for prediction. 

DMD approximates the dynamics of a high dimensional, and potentially nonlinear, system with a linear model that advances the system state forward in time \cite{schmid_dynamic_2010,Rowley2009jfm}. 
Although DMD was originally developed to study fluid systems~\cite{schmid_dynamic_2010,schmid2011applications}, it has since been applied to a wide range of physics systems, such as neuroscience~\cite{brunton2016extracting,kunert2019extracting}, and it has also been rigorously connected to nonlinear systems through Koopman operator theory~\cite{Rowley2009jfm,hirsh2021structured,Brunton2021koopman}. 
Let us consider $\xv(t) \in \mathbb{R}^n$ as the state vector of an $n$-dimensional dynamical system at time $t$. 
In a discrete-time representation with uniform interval $\Delta t$, we can denote the system state at each time step with subscript $k \in \mathbb{Z}$ such as $\xv_k = \xv(k\Delta t)$. 
Now consider $T$ time samples of the state, i.e., $k\in\{1,...,T\}$, and define two state snapshot matrices  $\Xv = [\xv_1 \ ... \ \xv_{T-1}]$ and $\Xv' = [\xv_2 \ ... \ \xv_T]$. The goal of DMD is to compute a state transition matrix, $\Av\in\mathbb{R}^{n\times n}$, that propagates the data forward by one time step, i.e., $\Av$ such that $\Xv' = \Av\Xv$. To do so, DMD uses a singular value decomposition (SVD) to compute the pseudo-inverse of the data matrix $\Xv$, which has the added benefit of allowing a reduced-order approximation to $\Xv$ based on how many modes, or singular values, are kept. Thus, $\Av$ provides a linear approximation of the system dynamics.

The process of calculating $\Av$ can be broken into the four steps outlined below \cite{kutz_dynamic_2016,Brunton2019book}:
\begin{enumerate}
    \item Perform SVD on $\Xv$: 
    \begin{equation*}
        \Xv = \Uv\Sigv \Vv^*,
    \end{equation*} 
    where $\Vv^*$ denotes the complex conjugate transpose of $\Vv$.
    
    \item Project $\Av$ onto the first $r$ most dominant modes: 
    \begin{equation*}
        %\Xv' = \Av \Uv_r \Sigv_r\Vv^*_r \Rightarrow 
        \tilde{\Av} = \Uv^*_r \Av \Uv_r = \Uv^*_r \Xv' \Vv_r \Sigv^{-1}_r,
    \end{equation*}
    where $\Uv_r \in \mathbb{R}^{n\times r}$, $\Vv_r \in\mathbb{R}^{(T-1)\times r}$, and $\Sigv_r\in\mathbb{R}^{r\times r}$ are the $r$-truncated singular vector and value matrices.
    
    \item Compute the eigenvalues and eigenvectors of $\tilde{\Av}$: \begin{equation*}
        \tilde{\Av}\tilde{\Wv} = \tilde{\Wv}\Lamv.
    \end{equation*}
    The eigenvalues of $\tilde{\Av}$ are also eigenvalues of $\Av$.
    
    \item Compute the eigenvectors of $\Av$: \begin{equation*}
        \Wv = \Xv'\Vv_r\Sigv^{-1}_r\tilde{\Wv}.
    \end{equation*}
\end{enumerate}
With the eigenvalues and eigenvectors of $\Av$, the future state can be predicted. 
Knowing $\Av = \Wv \Lamv \Wv^*$, 
and that $\Av$ provides a one time step forward approximation, we can write $\xv_{k+1} = \Av \xv_k$ and, we can write $\xv_{k}$ as:
\begin{align}
    \xv_{k+1} \approx \Wv e^{k \Delta t \Lamv} \Wv^* \xv_1 \notag.
\end{align}

By projecting onto the dominant modes in Step 2, DMD provides a computationally efficient way to compute the transition matrix $\Av$, thus making it a useful method when dealing with high-dimensional dynamics~\cite{Taira2017aiaa,Taira2020aiaaj}, such as for multi-agent systems. 
The dominant modes are those singular vector groups with the largest magnitude singular values, indicating that those corresponding singular vectors have the most influence on the structure of the matrix. 
By choosing to keep only the most dominant modes, one still retains the key properties of the matrix, without the computational burden of performing an eigen-decomposition on a full-rank matrix. 

Standard DMD does not explicitly model the influence of control inputs.
However, it is often convenient to represent the dynamics of an individual swarm agent using a state-space model with a control input that is dependent on the states of its neighboring agents.
DMD with control (DMDc)~\cite{proctor_dynamic_2016} captures this structure and is used as the base for the development of swarmDMD. DMDc is used to determine the state transition matrix, $\Av$, and the control matrix, $\Bv$, of a linear system approximation of the form:
\begin{align}
    \xv_{k+1} = \Av \xv_k + \Bv \uv_k, \label{eq:ControlSystem}
\end{align}
where $\xv_k$ is the system state and $\uv_k$ is the control input at time step $k$. 
Depending on whether $\Bv$ is known, DMDc can be used to learn either just the $\Av$ matrix or both $\Av$ and $\Bv$. 
As we will show in Section~\ref{sec:swarmDMD}, \eqref{eq:ControlSystem} motivates the development of swarmDMD. 

\section{DMD for Swarms} \label{sec:swarmDMD}

The swarmDMD algorithm is designed to use data we can observe/measure from the swarm, such as agent position, velocity, and inter-agent distance, to create an augmented ``state" vector, which is used as the control input to an agent's position dynamics. 
We assume that the coupled multi-agent dynamics can be considered control-affine. The goal is to determine the feedback control matrix, which determines the inter-agent interactions. 

Consider a swarm of $N$ agents.
We define the 2D location of agent-$i$ at time $t$ as $\mathbf{p}^i_t = \mathbf{p}^i(t) \triangleq [x^i(t), y^i(t)]\T$ for $i\in (1, 2, \dots, N)$, the velocity as $\vv^i_t = \vv^i(t) \triangleq [v_x^i(t), v_y^i(t)]\T$, and the heading as $\theta_t^i = \theta^i(t)$. 
The state of agent-$i$ at time $t$ is denoted by $\xv^i_t$, which could be comprised of any of the aforementioned variables. 
By stacking the states of every agent we obtain the state vector $\xv_t=[(\xv^1_t)\T,...,(\xv^N_t)\T]\T \in\mathbb{R}^{Nw}$, where $w$ is the number of state variables chosen for each agent. 
For any variable written without a superscript, it is assumed that this variable has been formed by concatenating over all agents. 
We consider discrete-time systems, with $t=k\Delta t$ for some time step $\Delta t$, and simplify the notation of time index to a subscript $k$. 

\begin{figure*}
    \centering
    \includegraphics[width=0.88\linewidth]{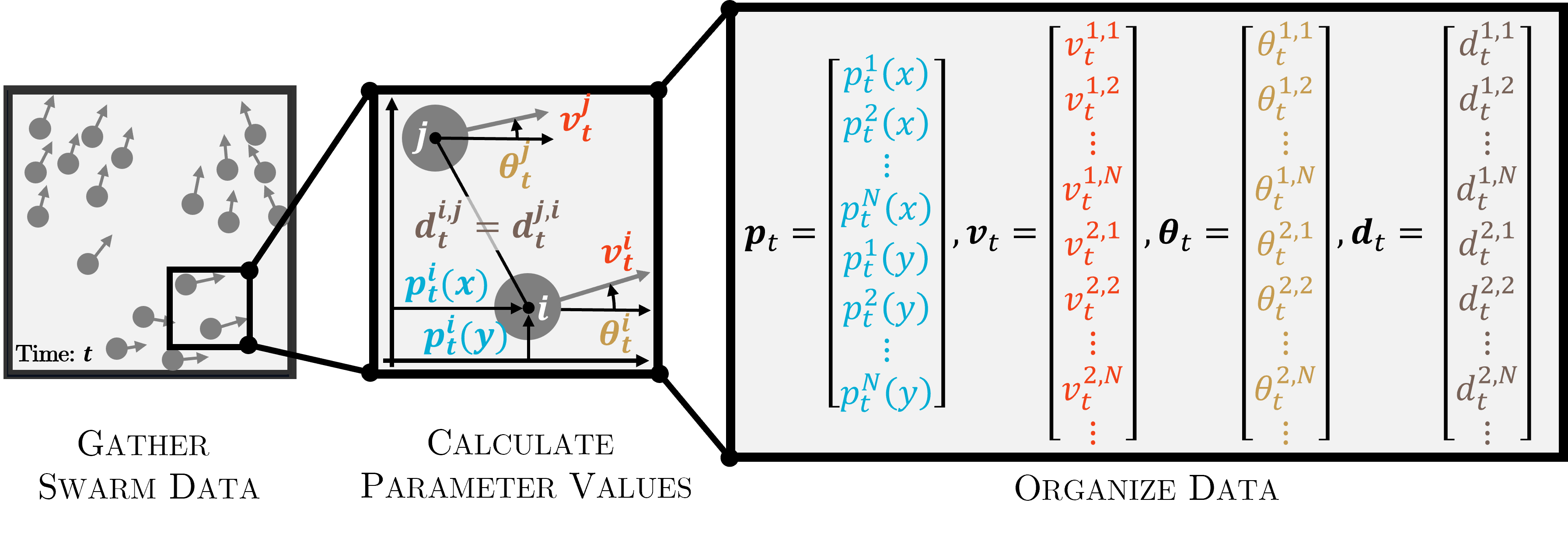}
    \includegraphics[width=0.88\linewidth]{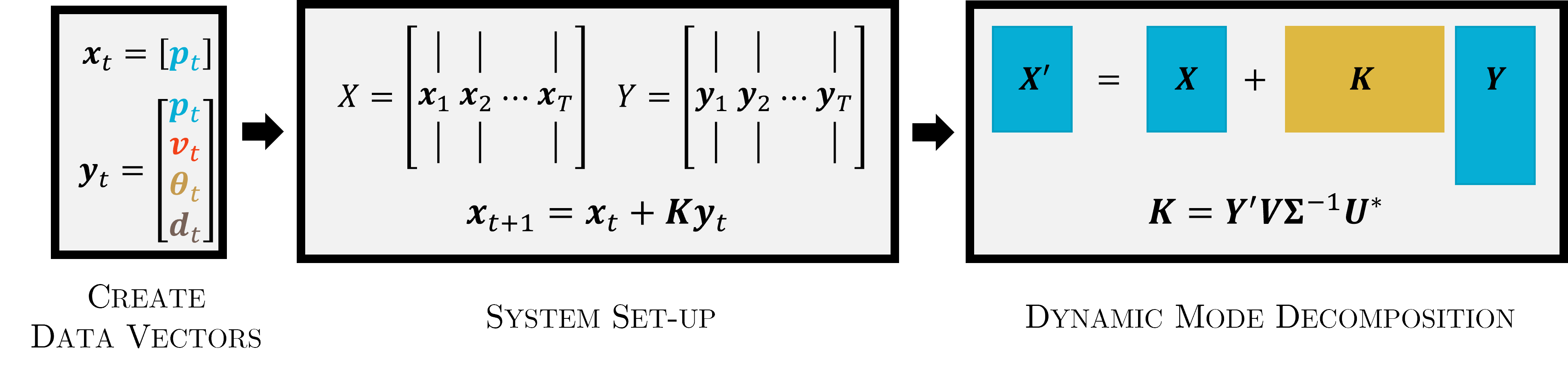}
    \caption{Illustration of the swarmDMD process for 2D swarm data, from data collection to the computation of the control matrix $\Kv$. Data is first collected from the desired swarm, in this case position information (blue), and is then used to calculate other desired properties such as velocity (orange), heading (yellow), and inter-agent distance (brown). We then take the state, $\xv$, to be position, and the control, $\yv$, to be the concatenation of position, velocity, heading, and inter-agent distance. This is then used to calculate the feedback control matrix $\Kv$.}
    \label{fig:SummaryFigure}
\end{figure*}

Recall from the system dynamics in \eqref{eq:ControlSystem} that we must learn the forcing term $\Bv\uv_k$.
Assuming the agents operate under a feedback-type control law, we write $\uv_k^i = \Fv^i \yv_k^i$, where $\yv_k^i$ is an augmented state vector used to determine the control input, and $\Fv^i$ is the feedback matrix for agent-$i$. 
This augmented state vector $\yv_k^i\in\mathbb{R}^m$ can include measurement data of individual agents, such as agent position,  velocity, and heading as defined before, as well as derived quantities such as relative position $\Delta \pv_k^{ij} = |\pv_k^i-\pv_k^j|$, relative distance $d^{ij}_k = ||\pv_k^i-\pv_k^j||_2$, relative heading $\Delta \theta^{ij}_k = |\theta_k^i-\theta_k^j|$,  relative velocity $\Delta \vv_k^{ij} = |\vv_k^i-\vv_k^j|$, and relative speed $\Delta v_k^{ij} = \left| ||\vv_k^i||_2 - ||\vv_k^j||_2 \right| $. 
By stacking the control vectors we obtain $\yv_k = [(\yv_k^1)\T,...,(\yv_k^N)\T]\T\in\mathbb{R}^{Nm}$. Thus, \eqref{eq:ControlSystem} may be rewritten as:
\begin{align}
    \xv_{k+1} = \xv_k + \Kv \yv_k,
    \label{eq:SystemSetup}
\end{align}
where $\Kv\in\mathbb{R}^{Nw\times Nm}$, and $\Kv = \Bv\Fv$. 
Note that we have fixed the state dynamics to be the identity, $\Av=\mathbf{I}$, so that all interactions influencing an agent are considered as external inputs and don't depend on its own internal state. 
Rearranging \eqref{eq:SystemSetup} yields
\begin{align}
    \xv_{k+1}-\xv_k = \Kv \yv_k,
\end{align}
resulting in a system that resembles the original DMD regression. Thus, DMD can be used to determine $\Kv$. 
As in standard DMD, $T$ snapshots are arranged into matrices $\Xv = [\xv_1 \ ... \ \xv_{T-1}]$, $\Xv' = [\xv_2 \ ... \ \xv_T] $ and $\Yv = [\yv_1 \ ... \ \yv_{T-1}]$. 
Equation (\ref{eq:SystemSetup}) may be written in matrix form as
\begin{align}
    \Xv' - \Xv = \Kv \Yv. \notag
\end{align}
By defining $\Sv = \Xv'-\Xv$, we arrive at a linear approximation of the swarm dynamics
\begin{equation}\label{eq:SwarmDMD}
    \Sv = \Kv \Yv. 
\end{equation}
Taking the rank-$r$ SVD of $\Yv$ to obtain $\Uv\in\mathbb{R}^{Nm\times r}, \mathbf{\Sigma}\in\mathbb{R}^{r\times r}$, and  $\Vv\in\mathbb{R}^{(T-1)\times r}$, it is possible to approximate $\Kv$ as: 
\begin{align}
    \Kv \approx \Sv \Vv \mathbf{\Sigma}^{-1} \Uv^*, \label{eq:K}
\end{align}
which can be used to predict future dynamics as:
\begin{align}
    \xv_{k+1} = \xv_k + \Kv \yv_k. \notag
\end{align}
The swarmDMD process, from data collection to the computation of $\Kv$, is outlined in Figure \ref{fig:SummaryFigure}.

We note that \eqref{eq:SwarmDMD} is different from the conventional DMD formulation in that $\Sv$ may have a different shape than $\Yv$, leading to a rectangular $\Kv$. 
There are several reasons for this formulation, as opposed to simply stacking the desired data (velocity, heading, inter-agent distance, etc.) in the state vector and using standard DMD. 
First, the swarmDMD formulation yields a particular interpretation, as discussed below. 
Further, it can be shown that for any combination of data in swarmDMD that includes at least one data type that is not position, the computational complexity of swarmDMD is less than that of a traditional stacked DMD. 
Moreover, stacked DMD would seek a model for the evolution of all of the variables, which is known to have closure issues~\cite{Brunton2021koopman}. 

Note that additional regularizers $R(\cdot)$ may be included in the swarmDMD optimization problem
\begin{align}
    \argmin_{\Kv} \|\Sv-\Kv\Yv\| + \lambda R(\Kv).
\end{align}
For example, sparsity may be useful to promote an influence matrix $\Kv$ where each agent uses minimal information in its control law. 
Similarly, group sparsity may be used to promote a group of agents having similar influence laws.  
Both of these are promising avenues of ongoing research.  

\subsection{Interpreting $\Kv$}
To interpret the swarm influence matrix $\Kv$, it is important to note how the variables are grouped in the vector $\yv$.  Each variable type, such as the $x$ or $y$ component of the agent position, are grouped together across all agents, and then these variable groups are concatenated as in Figure~\ref{fig:SummaryFigure}.  

If the state corresponds to the agents' positions, then $w=2$. From the definition of $\Kv$ in \eqref{eq:K}, recall $\Sv = \Xv'-\Xv$ a matrix of component-wise speeds of each agent at each time step. 
The rows of $\Sv$ contain, for each agent $i$, the $x$ and $y$-components of the agent's speed,  $\sv\T_{i,x}$ and $\sv\T_{i,y}$ respectively,  for $i\in\{1,..,N\}$. 
Define the concatenated vector as $\sv^T_i$ and let
\begin{align}
\begin{split}
    &\Sv = \left[\begin{array}{c}
        \sv\T_{1}\\
        \vdots \\
         \sv_{2N}\T  
    \end{array} \right], \ \Vv = \left[\begin{array}{ccc}
        \vv_1 & \cdots & \vv_r  
    \end{array} \right], \\
    &\Sigv = \diag(\sigma_1, ..., \sigma_r), \ \Uv = \left[\begin{array}{c}
        \uv_1\T\\
         \vdots \\
         \uv_{Nm}\T
    \end{array} \right].
\end{split}
\end{align}
The individual entries, $k_{i,j}$, of $\Kv$ can be written as:\vspace{-0.2cm}
\begin{align}
    k_{i,j} = \sum_{\ell =1}^r \sigma_\ell^{-1} u_{\ell j} \sv_{i}\T \vv_\ell = \bar{\uv}_{j}\T \Sigv^{-1} \Vv\T \sv_{i}
\end{align}
where $\bar{\uv}\T_{j}\in\mathbb{R}^{r}$ is the $j$-th row of $\Uv$. Each entry $k_{i,j}$ of $\Kv$ is a projection of the speed onto the $j$-th component of each column of $\Uv$. Since the columns of $\Uv$ are left singular vectors correspond to modes, and these modes are ordered from most influential to least influential, then each subsequent entry in $\bar{\uv}_{j}\T$ carries less influence than the last. 
Thus, each row of $\Kv$, which determines the combination of data that each agent receives for its $x$- and $y$-positions, is given by:
\begin{align}
    \kv_{i} = \left[\begin{array}{ccc}
        \bar{\uv}_{1}\T \Sigv^{-1} \Vv\T \sv_{i} & \cdots & \bar{\uv}_{Nm}\T \Sigv^{-1} \Vv\T \sv_{i} 
    \end{array} \right] . \label{eq:K row}
\end{align}
Recall $\sv_{i}\T$ is the vector of speed snapshots at each time step and $\Vv$ encodes the temporal correlation within the snapshot data. Starting with $\Vv\T \sv_{i}$, the columns $\vv_k$ of $\Vv$ indicate which instants are most influential to the change of state-$i$. This influence is scaled by $\Sigv^{-1}$. Finally, multiplying with $\bar{\uv}_j$ determines which components $j$ will have the greatest influence on component $i$, looking over the entire row $i$ in (\ref{eq:K row}). In a sense, $\Kv$ selects the states, neighbors, and time instants that are most influential to an agent's position change, acting as a control matrix that encodes the multi-agent interaction laws inherent to the observation data. 

\subsection{Alternative Dynamics Formulations}
In addition to the dynamics given in \eqref{eq:SystemSetup}, which we will call the standard dynamics, we also considered implementations which assumed first-order Cartesian and polar dynamics. The first-order Cartesian dynamics formulation is set up as
\begin{align}\label{Eq:FO:Cartesian}
    \xv_{k+1} = \xv_k + \vv_k \Delta t + \Kv \yv_k,
\end{align}
where
\begin{align}
    \xv_k = \pv_k, \;\yv_k =  \left[\begin{array}{c}
        \Delta \pv_k \\
        \Delta \vv_k 
    \end{array} \right], \notag
\end{align}
and $\vv_k$ is defined as before.
Similarly, the first-order polar dynamics formulation is set up as
\begin{align}\label{Eq:FO:Polar}
    \xv_{k+1} = \xv_k +  \left[\begin{array}{c}
        v_k \cos \theta_k\\
        v_k \sin \theta_k
    \end{array} \right] \Delta t + \Kv \yv_k,
\end{align}
where
\begin{align}
    \xv_k =  \pv_k \textnormal{ and } \yv_k = \left[\begin{array}{c}
        \Delta d_k \\
        \Delta v_k \\
        \Delta \theta_k 
        \end{array} \right]. \notag
\end{align}
These alternative formulations provide flexibility in capturing various types of swarm behavior and dynamics.  

\section{Experiment Set-up} \label{sec:setup}
In this section, we introduce the swarm simulation and the metrics that were used to assess the performance of swarmDMD.  Swarm data is generated from the flocking behaviour following the Vicsek model. We provide information about the swarmDMD set-up, the ground truth models used for training, and the metrics used in evaluating the results. 

% For the control input, the standard implementation of swarmDMD uses relative inter-agent distance data, calculated from the absolute positions. In the first order Cartesian method the data used for the control input is relative position and relative velocity, and the first order polar method uses relative distance, speed, and heading. That is,
% %
% \begin{align*}
%     \textnormal{Std.: } \yv = \left[ \Delta \pv  \right], 
%     \textnormal{ Cart.: } \yv =  \left[\begin{array}{c}
%         \Delta \pv \\
%         \Delta \vv
%     \end{array} \right], 
%     \textnormal{ Polar: } \yv = \left[\begin{array}{c}
%         \Delta d \\
%         \Delta v \\
%         \Delta \theta 
%         \end{array} \right].
% \end{align*}
%

\subsection{Vicsek Model} \label{sec:vicsek}
We use the Vicsek swarm model to generate data for training and analysis. The Vicsek model is a simple and well-studied model that produces flocking behaviour similar to that exhibited in biological swarm systems \cite{vicsek_novel_1995}. 
In the original Vicsek model, agents possess a constant forward speed and interact with the swarm by aligning their heading direction with the average heading of their neighbors in a certain Euclidean radius. 

The agent dynamics of the Vicsek model follow:
\begin{align}
    \xv^i_{k+1} = \xv^i_{k} + \vv^i_{k} \Delta t, \label{eq:Vicsek}
\end{align}
where $\vv^i_{k}$ is the velocity of agent-$i$, which acts like a control input to the linear system, and $\Delta t$ is the length of the time step. The velocity at time step $k$ is determined from the heading angle $\theta^i_k$ and the constant speed $\nu$, where $\theta^i_k = \langle \thetav_{k-1}\rangle_r+ \hat{\theta}^i$, and $\langle \thetav_{k-1}\rangle_r$ is the average direction of the agents, including agent-$i$, in a radius $r$ about agent-$i$, formally defined as:
\begin{align}
\langle \thetav_{k-1}\rangle_r = \arctan(\langle\sin\thetav_{k-1}\rangle_r/\langle \cos\thetav_{k-1}\rangle_r). \notag
\end{align}
Here, $\hat{\theta}^i$ is a random heading perturbation chosen with uniform probability from the interval $[-\eta/2,\eta/2]$. The values of the interaction radius, $r$, and randomness, $\eta$, characterize the emergent behaviour of a Vicsek swarm.

In addition to conventional flocking behaviours, we are also interested in how swarmDMD performs on systems with more complicated dynamics, such as milling. As a result, we also consider the modification to the original Vicsek model presented by Costanzo and Hemelrijk \cite{costanzo_spontaneous_2018}. In this modification, the field of view of each agent is restricted and a bound is placed on the agents' angular velocity, such that
\begin{align}
    \theta^i_{k} = \left\lbrace \begin{array}{ll}
        \langle\thetav_{k-1}\rangle_{r,\phi} + \hat{\theta}^i & |\Delta\theta^i| < \omega \Delta t   \\
        \theta^i_{k-1} + \omega\Delta t + \hat{\theta}^i & \Delta\theta^i \geq \omega \Delta t \\
        \theta^i_{k-1} - \omega\Delta t + \hat{\theta}^i & \Delta\theta^i \leq -\omega\Delta t
    \end{array} \right..
\end{align}
Here, $\langle\thetav_{k-1}\rangle_{r,\phi}$ is the average agent heading, as before, except now only agents in the field of view $\phi\in[0,2\pi]$ of agent-$i$ are considered, $\omega\in[0, \pi/\Delta t]$ is the maximal angular velocity, and $\Delta \theta^i\in[-\pi,\pi]$ is the difference between the current orientation and the average orientation. Please see \cite{costanzo_spontaneous_2018} for precise definitions of the field of view and difference in orientation. 

The parameter values used in the ground truth simulations are given in Table \ref{tab:GT_parameters}. A time step of 0.1s is used in the standard Vicsek model simulation and the swarmDMD recreation, and a time step of $1$s is used in the modified Vicsek model for milling. The training period is 5s and the prediction period is 5s post-training. Before being used in swarmDMD, the milling swarm data was interpolated to have a time step of $0.1$s, and the number of agents was reduced to 200 via random sampling. We use the first eight most dominant DMD modes in our swarmDMD recreation and prediction. 
\begin{table}[]
    \caption{Parameters used in the Vicsek models for ground truth simulations. Parameters are compiled from \cite{vicsek_novel_1995,costanzo_spontaneous_2018}. }
    \centering
    \begin{tabular}{c|c c}
        \hline
        \hline
         & Standard & Milling\\
         \hline
        $N$ & 50 & 1000 \\
        $\Delta t$ & 0.1 & 1 \\
        $\rho$ & 16 & 2.5 \\
        $r$ & 0.05, 0.25, 0.5 & 1 \\
        $\phi$ & N/A  & $\pi/2$  \\
        $\omega$ & N/A & $\pi/18$ \\
        $\eta$ & 0, $\pi/12$ & $0.5 \omega/\Delta t$ \\
        $\nu$ & 0.03 & $1.03 r \omega$\\
        \hline
        \hline
    \end{tabular}
    \label{tab:GT_parameters}
\end{table}

\begin{figure*}
    \centering
    \vspace{-.1in}
    \includegraphics[width=0.88\linewidth]{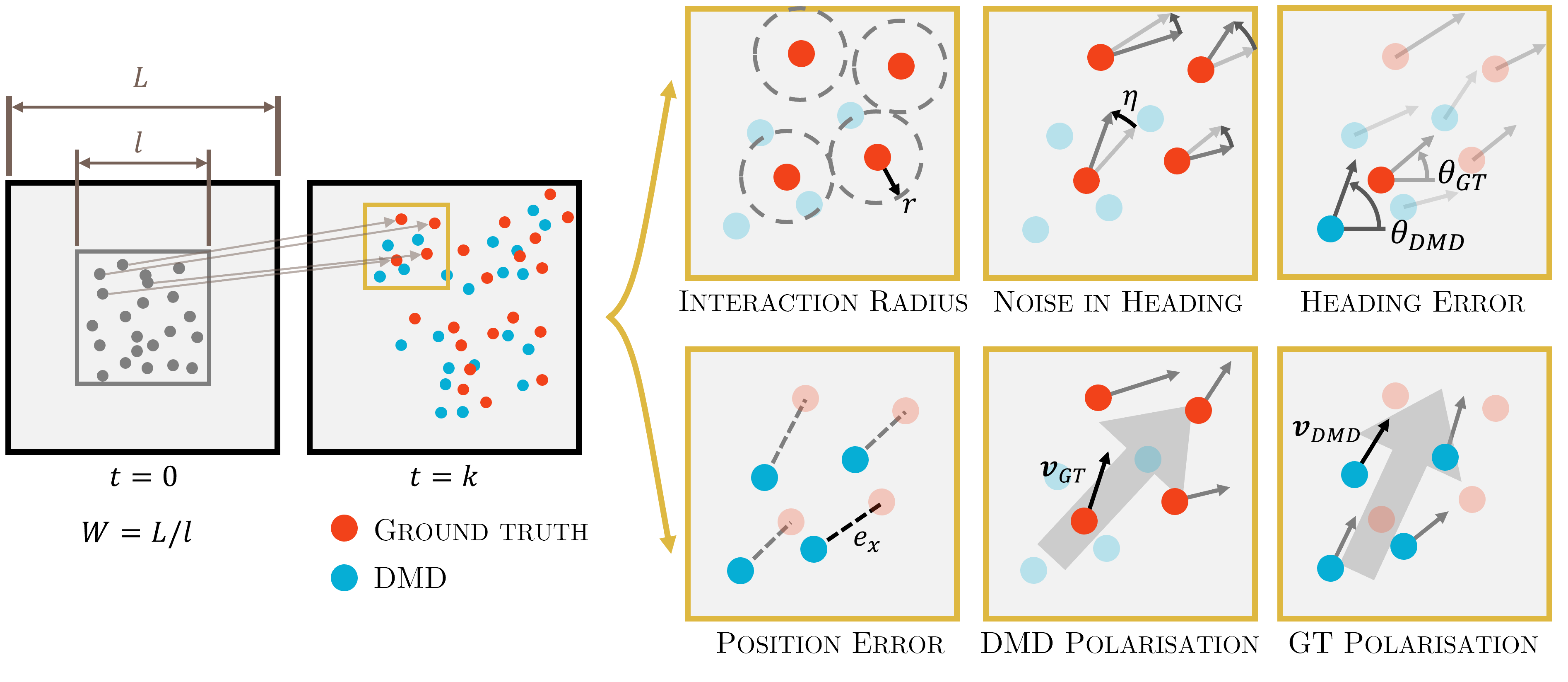}
    \vspace{-.1in}
    \caption{A visual representation of key parameters used in simulations: domain size, interaction radius, noise in heading; and four of the metrics used to analyse the results: heading error, position error, and polarisation. In the first frame at $t=0$, swarmDMD is initialzed with the ground truth agent positions. 
    In a future frame ($t = k$) when positions predicted by swarmDMD have diverged from the ground truth, the ground truth agent position is shown in orange, and the swarmDMD prediction in blue. Here, $L$ is the width of the square simulation domain, and $l$ the width of the square domain within which agents are initialized following a uniform distribution. The large grey arrows in the background of the Polarisation frames indicate the average direction of the swarm. }
    \label{fig:metrics_summary}
\end{figure*}

\subsection{Evaluation Metrics} \label{sec:metrics}

We now discuss the metrics used to evaluate reconstruction and prediction. The metrics we consider are: position error, heading error, polarisation error, error in angular momentum, and agent distribution about a focal agent. The first three metrics and the key simulation parameters are shown in Figure \ref{fig:metrics_summary}.

The position and heading errors are direct comparisons between the absolute position measurements from the ground truth and swarmDMD reconstruction. An agent's position error is the distance between the reconstructed and actual position. Similarly, an agent's heading error is defined as the absolute heading difference between the swarmDMD reconstruction and the ground truth. These two errors are averaged over all agents.

% Described in this section are the metrics used to evaluate the resulting swarm reconstruction via the feedback matrix $\Kv$, determined by swarmDMD. All metrics are described below, but it was found that the position error, nearest neighbour, and polarisation provided the most useful information in evaluating performance. These metrics are shown pictorially in Figure \ref{fig:metrics_summary}.

% The direct comparisons we use are between the $x$ position, $y$ position, and speed of the agents. For these comparisons the difference in each of these quantities between the ground truth and reconstructed swarms were plotted and the time at which the average of the error passed above a certain threshold was recorded for comparison between reconstructions.

The calculations of polarisation and angular momentum are borrowed from \cite{bhaskar_analyzing_2019}. We are interested in these two metrics as polarisation and angular momentum describe the collective motion of the swarm, and the degree to which the swarm is acting as one cohesive unit. Polarisation is calculated as:
\begin{align}
    P_k &= \left|\left| \frac{\sum_{i=1}^N \vv^{i}_k}{\sum_{i=1}^N ||\vv^{i}_k||_2} \right|\right|_2 \in [0,1], \label{eq:polarisation}
\end{align}
and it provides a characterisation of the orientation of the swarm and takes values between $0$ and $1$. A value of $P_k$ close to $1$ means the majority of agents in the swarm are travelling in a similar direction, i.e., they have similar orientation. Angular momentum is calculated as:
\begin{align}
    \begin{split}
    M_k &= \left|\left| \frac{\sum_{i=1}^N \pv^{i}_k\times \vv^{i}_k}{\sum_{i=1}^N ||\pv^{i}_k||_2 \ ||\vv^{i}_k||_2} \right|\right|_2 \in[0,1], \label{eq:momentum}
    \end{split}
\end{align}
and it takes values between $0$ and $1$, describing the normalised momentum of the swarm. Angular momentum gives a notion of the rotational motion within a swarm. The polarisation and angular momentum errors are calculated as the absolute difference between the ground truth and swarmDMD results. 

\begin{figure}
    \centering
    \includegraphics[width=1\linewidth]{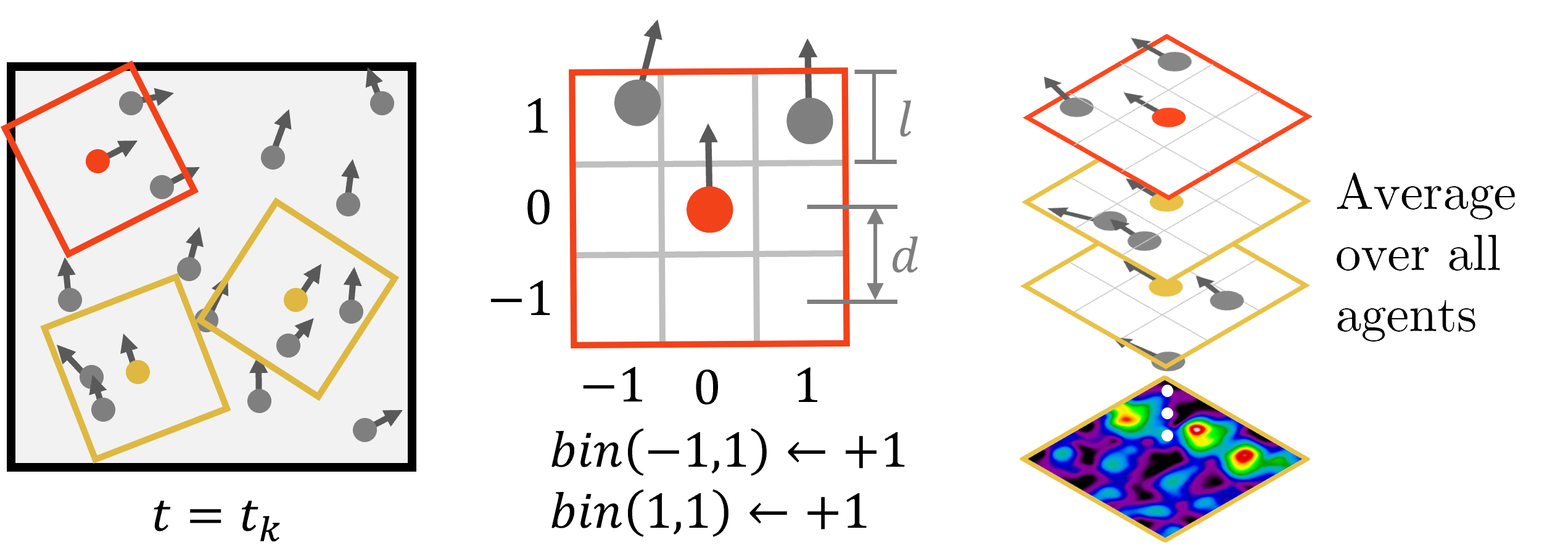}
    \caption{An illustration of the process for computing the density distribution of neighbours for each time step. The domain about each agent is discretised into bins, the number of neighbours in each bin is counted, converted to a percentage of the total number of agents in the neighbourhood, and then averaged over all agents in the swarm. The bins are distance $d$ apart and have width $l$. }
    \label{fig:density_illustration}
\end{figure}

The final metric considered is the distribution of neighbours, which helps to characterise the general structure of the swarm. Our distribution calculations are based on those in \cite{Katz2011}. To calculate the neighbour distribution for a single focal agent, a neighbourhood about the agent is divided into bins, and the number of neighbours in each bin is counted and divided by the total number of agents in the neighbourhood. Then, these bin counts are averaged over all agents in the swarm (one by one, treating each agent as a focal agent), and over the desired time frame. An illustration of the process for each time step $k$ is given in Figure~\ref{fig:density_illustration}. 

\section{Results}\label{sec:results}
We now discuss the performance of swarmDMD on several example scenarios and in two different implementation styles. The ground truth swarms in the first three scenarios are created from the standard Vicsek model, one scenario per interaction radius in Table \ref{tab:GT_parameters} and fixing $\eta=0$.
The fourth scenario uses the Vicsek milling model. The two implementation styles presented here both use the standard swarmDMD dynamics, but are initialised in different ways. The first implementation initialises the swarmDMD agents at the start time of the training period (termed Basic), whereas the second re-initialises the agents with the positions and headings of the ground truth agents every $g$ seconds, and propagates this for $h$ seconds (termed Re-initialisation); we take $g=0.5s$ and $h=10s$. The re-initialisation implementation was only applied to the standard Vicsek model scenarios. A short discussion of the performance of swarmDMD with the first-order Cartesian and polar dynamics is given at the end of the section, with the corresponding figures in the Appendix.
 
\begin{figure}
    \centering
    \begin{centering}
        \includegraphics[width=\linewidth]{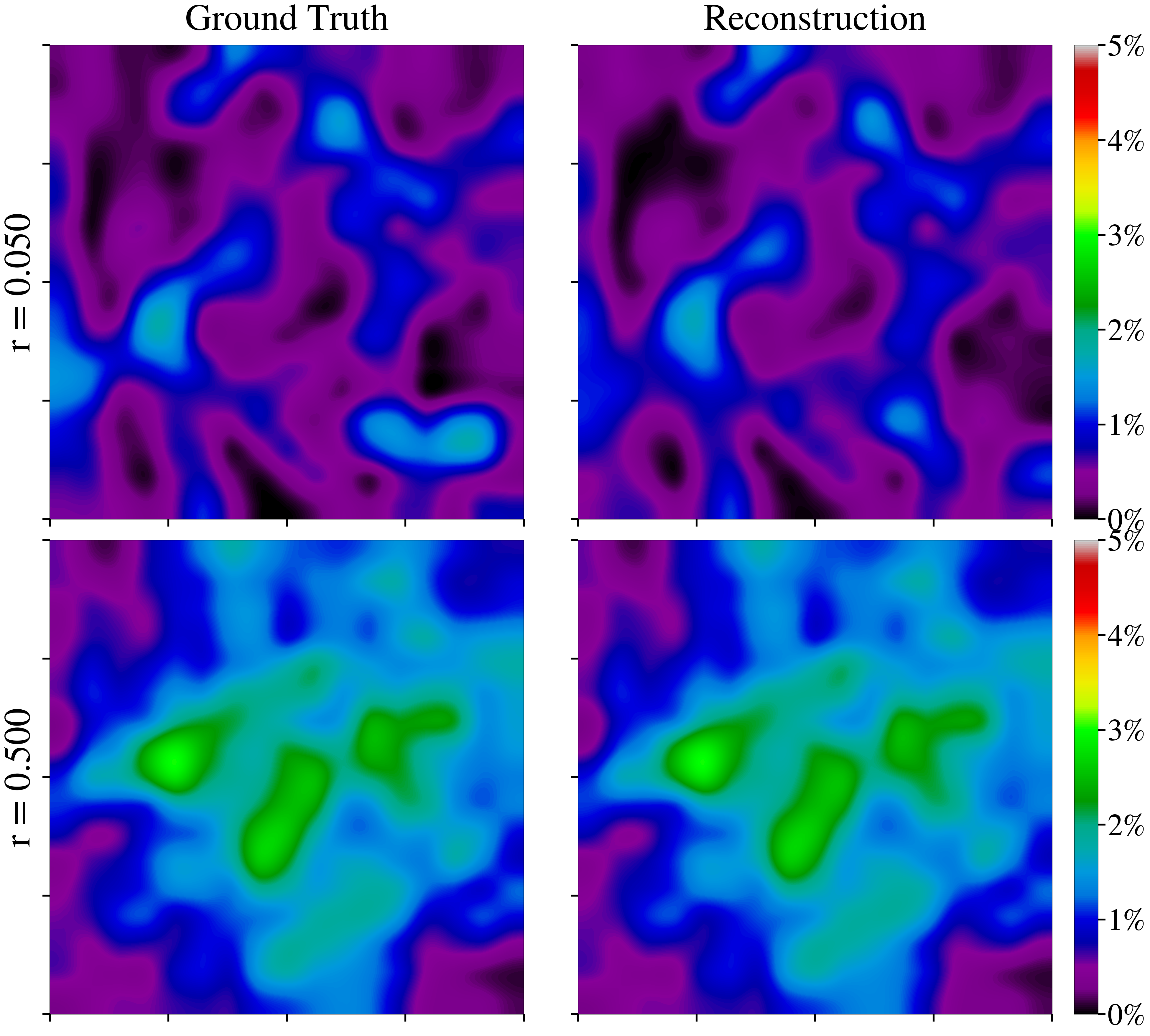} %
        \caption{Agent distribution analysis of swarmDMD with standard dynamics during the training period where the ground truth has $\eta=0$. On the left is the ground truth distribution, and on the right is the swarmDMD recreation. SwarmDMD captures the most prominent features in the case $r=0.05$, and is almost identical in the case $r=0.5$. }
        \label{fig:standard_densityT}
    \end{centering}
\end{figure}

\begin{figure}
    \centering
    \begin{centering}
    \includegraphics[width=\linewidth]{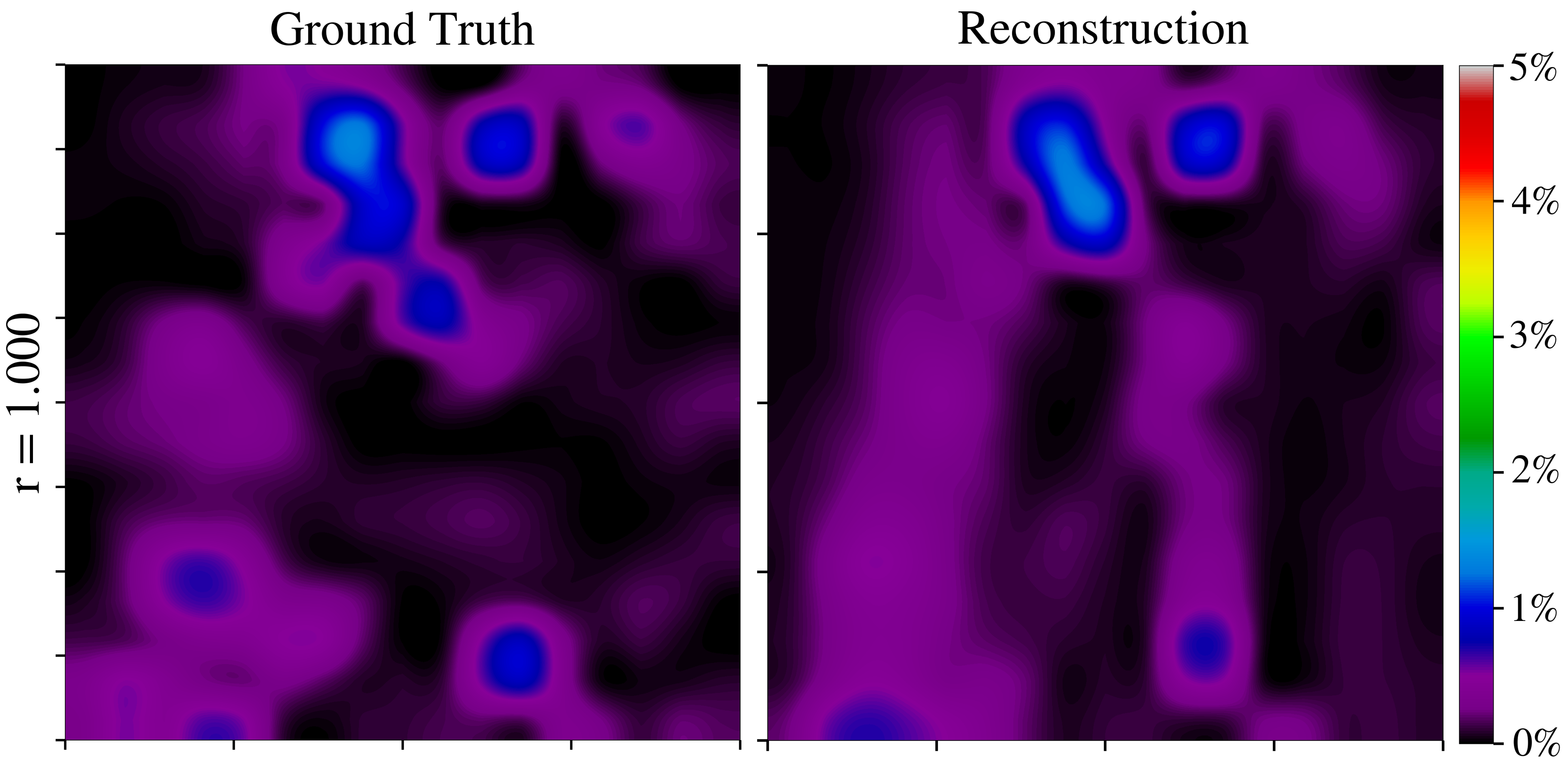}
    \caption{Agent distribution analysis of swarmDMD with standard dynamics during the training period where the ground truth is the milling case. On the left is the ground truth distribution, and on the right is the swarmDMD recreation. SwarmDMD does a decent job of capturing the ring-like pattern and three of the higher-likelihood areas.}
    \label{fig:milling_densityT}
    \end{centering}
\end{figure}

\begin{figure}
    \centering
    \includegraphics[width=\linewidth]{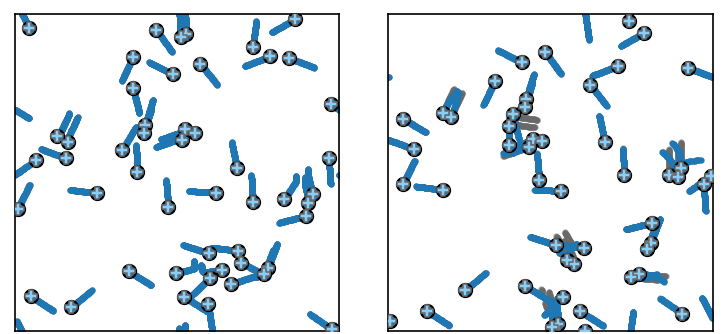}
    \includegraphics[width=\linewidth]{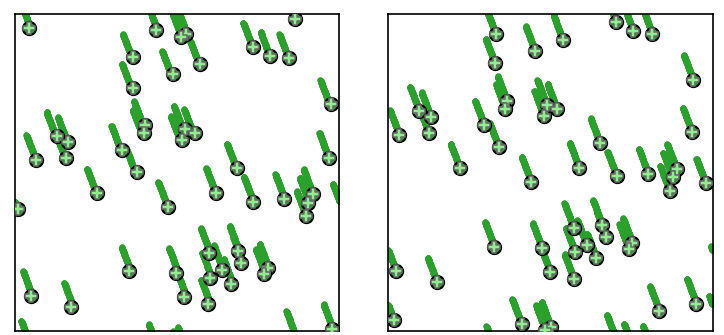}
    \includegraphics[width=\linewidth]{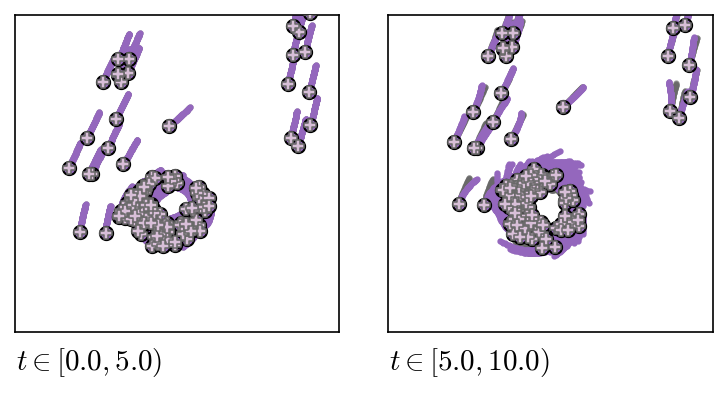}
    \caption{SwarmDMD agent trajectories for $r=0.05$ (blue), $r=0.5$ (green), and milling (purple). In all plots the ground truth trajectories are given in grey. The large circles outlined in black are the positions of the swarmDMD agents at the beginning of each time period, and the pluses outlined in light grey the positions of the ground truth agents; the lines leaving each of these indicate the trajectories of the agents over the specified time period.}
    \label{fig:trajectories}
\end{figure}

Figure \ref{fig:position} shows the swarmDMD reconstruction and prediction error in agents' positions. During the training period, the error is rarely above $10^{-4}$, demonstrating the ability of swarmDMD to accurately reconstruct the swarm dynamics. This is further reinforced by the agent density distribution shown in Figure \ref{fig:standard_densityT} for standard flock behaviour and Figure \ref{fig:milling_densityT} for milling, which indicate that the swarmDMD recreation captures the general structure of agents within the swarm. In addition, Figure \ref{fig:trajectories} plots the trajectories of the agents during the training period under different inter-agent interaction radii settings, and the difference between the ground truth and the swarmDMD recreation is imperceptible during the training period. It should also be noted that in Figure \ref{fig:position}, the error associated with the $r=0.5$ case is so small that it is not included in the plot.

\begin{figure}
    \centering
    \includegraphics[width=\linewidth]{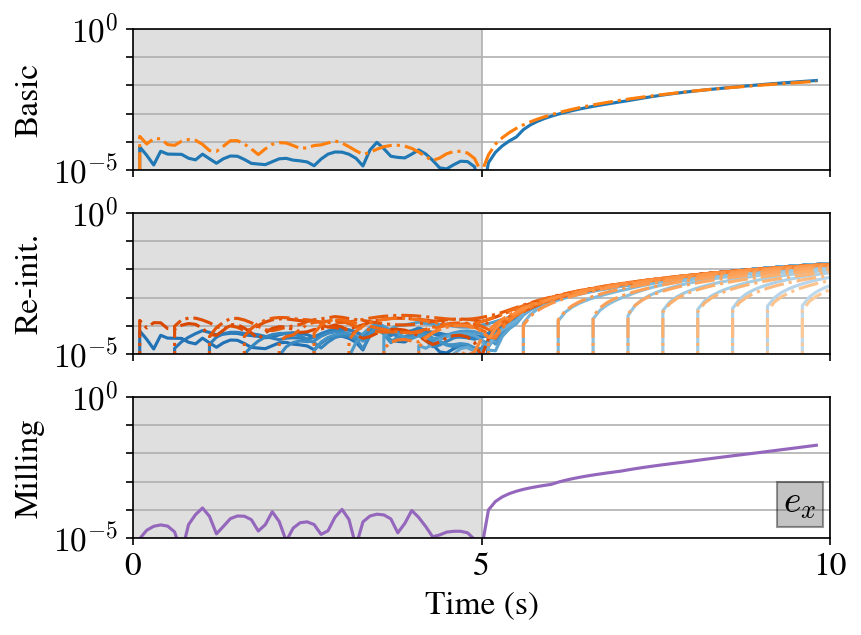}
    \caption{The resulting position error of the standard dynamics is compared across ground truth interaction radii, $r$, 0.05 (blue), 0.25 (orange), and 0.5 (green), and milling results (purple). Ground truth has $\eta=0$ for basic and re-initialisation. Three things are of note: (1) the low error during training across all interaction radii for both the basic and re-initialisation implementations, (2) the error is so small for $r=0.5$ that it is off the plot, and (3) initialising the algorithm outside the training period does not result in an increase in error.}
    \label{fig:position}
\end{figure}

\begin{figure}
    \centering
    \includegraphics[width=\linewidth]{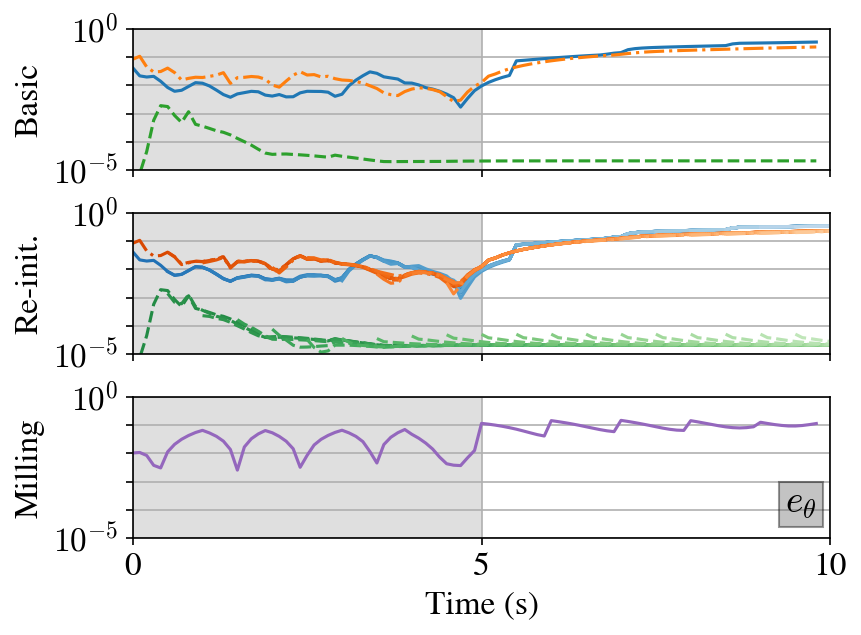}
    \caption{The resulting heading error of the standard dynamics is compared across ground truth interaction radii, $r$, 0.05 (blue), 0.25 (orange), and 0.5 (green), and milling results (purple). Ground truth has $\eta=0$ for basic and re-initialisation.}
    \label{fig:heading}
\end{figure}

\begin{figure}
    \centering    
    \begin{centering}
        \includegraphics[width=\linewidth]{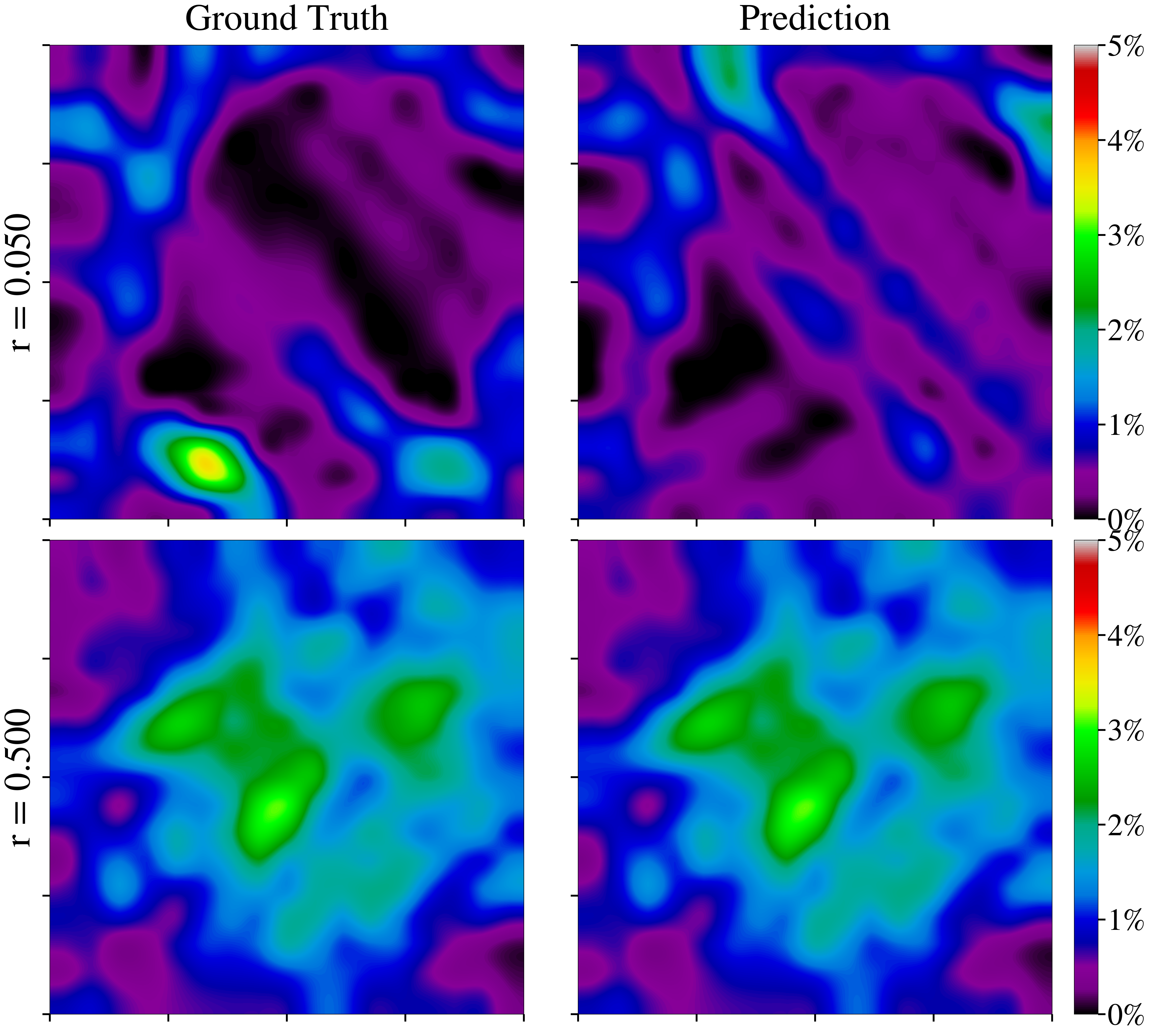}
        \caption{Agent distribution analysis of swarmDMD with standard dynamics over a period of 5s prediction, ground truth has $\eta=0$. On the left is the ground truth distribution, and on the right is the swarmDMD recreation. Again, swarmDMD captures the most prominent features (the diagonal stripe, and ring around the edges) in the case $r=0.05$, and is almost identical in the case $r=0.5$. }
        \label{fig:standard_densityP}
    \end{centering}
\end{figure}

\begin{figure}
    \centering
    \begin{centering}
    \includegraphics[width=\linewidth]{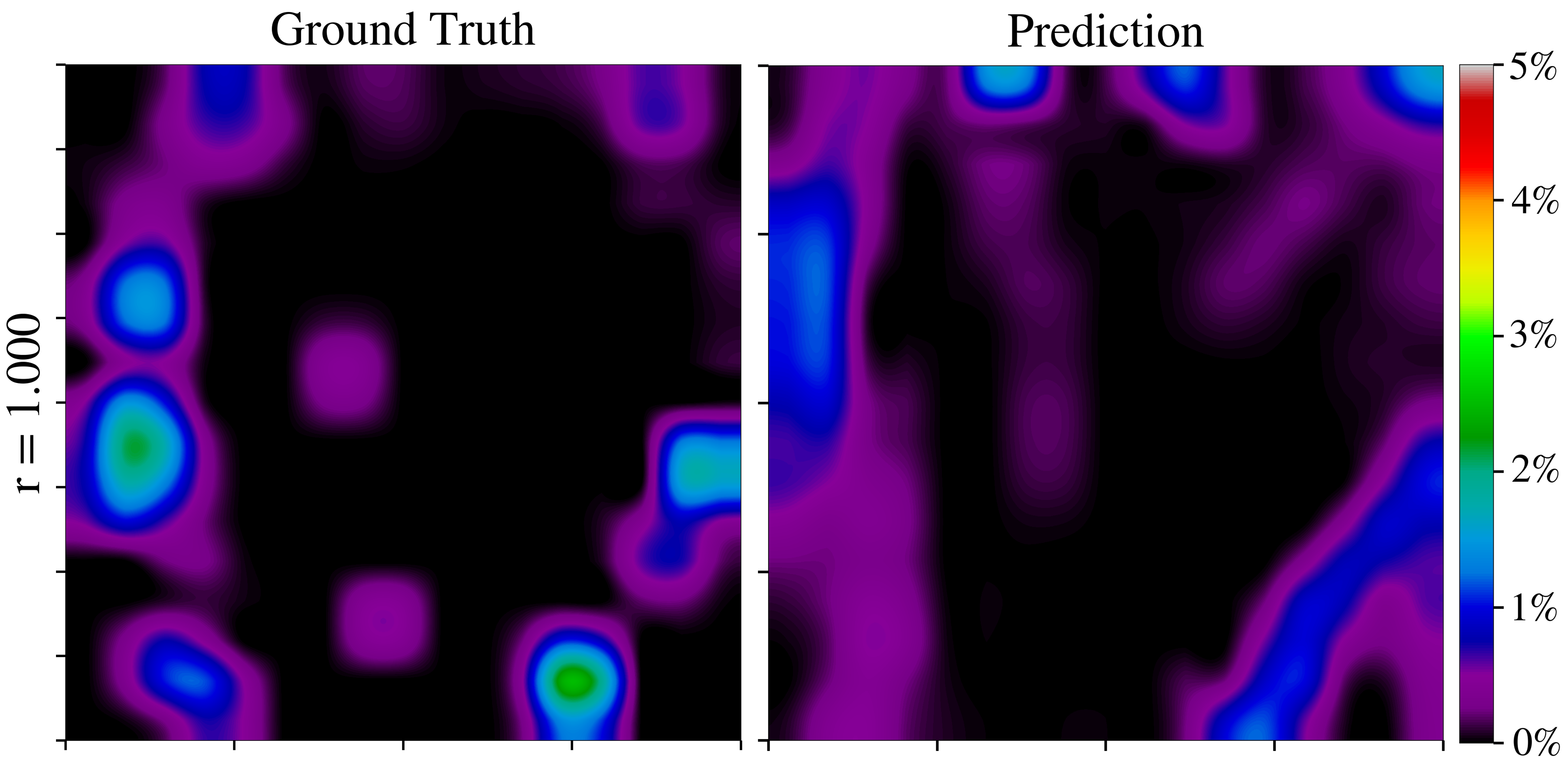}
    \caption{Agent distribution analysis of swarmDMD with standard dynamics over a period of 5s prediction, ground truth is the milling case. On the left is the ground truth distribution, and on the right is the swarmDMD recreation. It is interesting to note that here swarmDMD seems to capture the prominent patterns better than during the training period. (need to fix sizes of plots)}
    \label{fig:milling_densityP}
    \end{centering}
\end{figure}

Considering the post-training period of 5-10s, there is a window where the position error is still sufficiently small, indicating that swarmDMD can accurately predict agents' dynamics post-training for a short horizon. This is most evident in the position and heading error in Figures \ref{fig:position} and \ref{fig:heading}, but can also be seen to some extent across all error figures. In addition, the density distributions during prediction, given in Figures \ref{fig:standard_densityP} and \ref{fig:milling_densityP}, show that the general structure is captured by swarmDMD, though not as accurately as during training. This is a promising result that indicates the potential for swarmDMD models to be used for the control of swarms. It should be noted that of the ``basic" scenario plots are only included for $r=0.05$ and $r=0.5$, to avoid cluttering. It is encouraging that in all scenarios presented, swarmDMD captures the dominant distributions amongst agents during both training and prediction. 

\begin{figure}
    \centering
    \includegraphics[width=\linewidth]{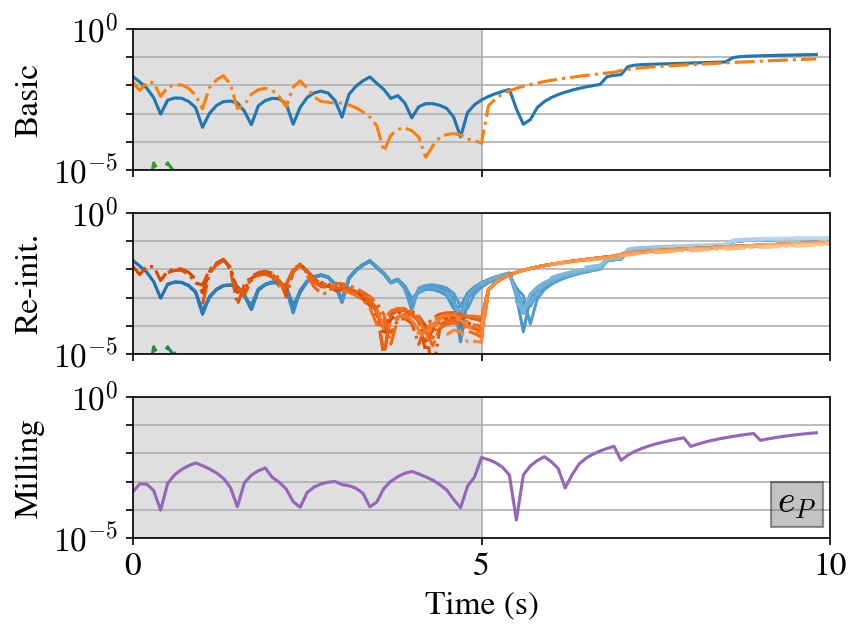}
    \caption{The resulting polarisation error of the standard dynamics is compared across ground truth interaction radii, $r$, 0.05 (blue), 0.25 (orange), and 0.5 (green), and milling results (purple). Ground truth has $\eta=0$ for basic and re-initialisation.}
    \label{fig:polarisation}
\end{figure}

\begin{figure}
    \centering
    \includegraphics[width=\linewidth]{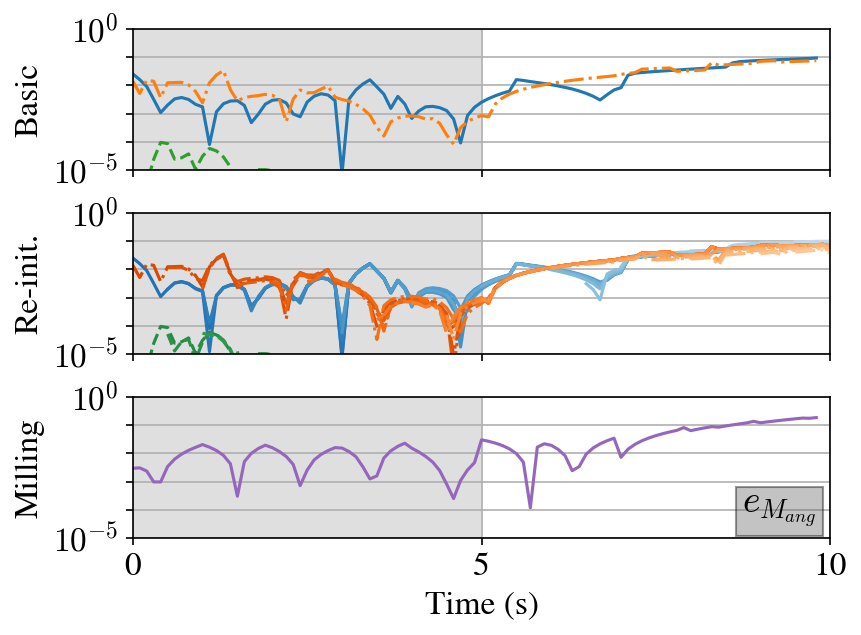}
    \caption{The resulting angular momentum error of the standard dynamics is compared across ground truth interaction radii, $r$, 0.05 (blue), 0.25 (orange), and 0.5 (green), and milling results (purple). Ground truth has $\eta=0$ for basic and re-initialisation.}
    \label{fig:momentum}
\end{figure}

The fact that the position error is very small for the milling scenario (Figure \ref{fig:position}) is significant; not only can swarmDMD recreate basic swarm flocking behaviours, it can reproduce milling motion and combined motions of milling and flocking in a single simulation. Figures \ref{fig:polarisation} and \ref{fig:momentum} also indicate that swarmDMD performs well with these more complicated dynamics, as the milling error is comparable to the ``basic" scenario errors with $r=0.25$. 
However, swarmDMD does have limitations, as seen in Figure \ref{fig:heading}, where the algorithm struggles with keeping the heading error low in the milling scenario. 
Importantly, by examining the re-initialisation results in Figures \ref{fig:position}, \ref{fig:heading}, and \ref{fig:momentum}, we observe that initialising the agents outside of the training period does not noticeably increase error, providing evidence of robustness in the model. 

Tables \ref{tab:errors-standard} and \ref{tab:times-standard} summarize the results shown before based on two metrics: the average error during the training period, and the amount of time the error stays below 1e-01 during the post-training prediction period. In these tables we also include the data from the scenarios with $\eta=\pi/12$, for which the results are shown in Figure \ref{fig:errors_noisy} in the Appendix. We can see that swarmDMD does perform better in the scenarios with less randomness in the motion of the agents. During training this is most noticeable in the position and heading, as the average polarisation and angular momentum errors during training for the scenarios with $\eta=\pi/12$ are of the same magnitude as those in the scenarios with $\eta=0$. This is expected, as the randomness in the agents' positions, if small enough, should only affect the motion of individual agents and not the general behaviour of the swarm.

\begin{table}[]
    \caption{Average error during training for the standard algorithm.}
    \centering
    \begin{tabular}{c c|c c c c}
         \hline
         \hline
        $r$ & $\eta$ & $e_x$ & $e_\theta$ & $e_P$ & $e_{M_{ang}}$ \\
        \hline
        0.05 & 0 & $2.95$e-05 & $1.03$e-02 & $4.11$e-03 & $3.69$e-03 \\
        0.25 & 0 & 7.07e-05 & 1.91e-02 & 4.74e-03 & 5.43e-03 \\
        0.5 & 0 & 7.60e-07 & 1.95e-04 & 1.45e-06 & 1.20e-05 \\
        \multicolumn{2}{c|}{Milling} & 3.67e-05 & 2.79e-02 & 1.21e-03 & 8.17e-03 \\
        0.05 & $\pi/12$ & 1.29e-04 & 5.79e-02 & 6.17e-03 & 8.04e-03 \\
        0.25 & $\pi/12$ & 1.49e-04 & 6.66e-01 & 6.63e-03 & 7.14e-03 \\
        0.5 & $\pi/12$ & 1.39e-04 & 6.17e-02 & 3.19e-03 & 5.44e-03 \\
         \hline
         \hline
    \end{tabular}
    \label{tab:errors-standard}
\end{table}

\begin{table}[]
    \caption{Length of time (in seconds) post-training that error remains below 1e-01 for the standard algorithm. The ``$-$" indicates the error is never larger than 1e-01 over the whole time period.}
    \centering
    \begin{tabular}{c c|c c c c}
         \hline
         \hline
         $r$ & $\eta$ & $t_x$ & $t_\theta$ & $t_P$ & $t_{M_{ang}}$ \\
        \hline
        0.05 & 0 & 12.7 & 1.3 & 3.7 & 5.3 \\
        0.25 & 0 & 18.1 & 1.6 & 5.8 & 6.7 \\
        0.5 & 0 & $-$ & $-$ & $-$ & $-$\\
        \multicolumn{2}{c|}{Milling} & 6.6 & 0 & 9 & 3.6 \\ 
        0.05 & $\pi/12$ & 5.7 & 0 & 2.4 & 3.3 \\
        0.25 & $\pi/12$ & 7.6 & 0 & 2.6 & 2.2 \\
        0.5 & $\pi/12$ & $-$ & 0.9 & $-$ & 5.3 \\
         \hline
         \hline
    \end{tabular}
    \label{tab:times-standard}
\end{table}

Tables \ref{tab:errors-FO} and \ref{tab:times-FO} give the average error during the training period, and the amount of time post-training for which the error stays below 1e-01, respectively, for the first-order Cartesian and polar dynamics, given by~\eqref{Eq:FO:Cartesian} and~\eqref{Eq:FO:Polar}, respectively. It is clear that the first-order Cartesian and polar dynamics do not perform as well as the standard implementation. Further discussion of these results can be found in the Appendix.

\begin{table}[]
    \caption{Average error during training for the Cartesian and polar algorithms with $\eta=0$.}
    \centering
    \begin{tabular}{c c|c c c c}
         \hline
         \hline
         & $r$ & $e_x$ & $e_\theta$ & $e_P$ & $e_{M_{ang}}$ \\
        \hline
        \multirow{3}{*}{Cart.} & 0.05 & 2.51e-04 & 1.31e-02 & 5.20e-03 & 4.57e-03 \\
        & 0.25 & 7.44e-02 & 5.55e-01 & 2.11e-01 & 1.94e-01 \\
        & 0.5  & 2.01e-01 & 1.13e+00 & 5.46e-01 & 4.88e-01\\
        \multirow{3}{*}{Polar} & 0.05 & 9.14e-04 & 3.51e-02 & 9.76e-03 & 8.64e-03 \\
        &  0.25  & 8.71e-03 & 2.11e-01 & 4.04e-02 & 5.69e-02 \\
        & 0.5 & 1.18e-03 & 2.41e-02 & 9.02e-03 & 8.51e-03 \\
         \hline
         \hline
    \end{tabular}
    \label{tab:errors-FO}
\end{table}

\begin{table}[]
    \caption{Length of time (in seconds) post-training that error remains below 1e-01 for the Cartesian and polar algorithms with $\eta=0$. }
    \centering
    \begin{tabular}{c c|c c c c}
         \hline
         \hline
         & $r$  & $t_x$ & $t_\theta$ & $t_P$ & $t_{M_{ang}}$ \\
        \hline
        \multirow{3}{*}{Cartesian} & 0.05 & 8.1 & 1 & 3.8 & 3.4 \\
        & 0.25 & 0 & 0 & 0 & 0 \\
        & 0.5  & 0 & 0 & 0 & 0\\
        \multirow{3}{*}{Polar} & 0.05 & 14 & 1.8 & 7.8 & 6.7 \\
        & 0.25  & 1.9 & 0 & 0 & 0 \\
        & 0.5 & 4.8 & 1.1 & 2.2 & 1.8 \\
         \hline
         \hline
    \end{tabular}
    \label{tab:times-FO}
\end{table}

\section{Conclusions}
% PARA1 - Summary of paper
% PARA2 - Limitations of the work/method (major assumptions, what it can't do)
% PARA3 - Future works
In this paper, we introduced for the first time the method of swarmDMD to learn local interaction laws that give rise to swarm motions from pure observation data. We demonstrated the performance of swarmDMD on both flocking and milling swarm motions generated with the Vicsek model and its variants. It was concluded that swarmDMD can not only reconstruct the swarm behaviour faithfully but also predict the swarm motion sufficiently accurately within a short window after the training period. We tested swarmDMD on different observation data types and found inter-agent distance to give the best reconstruction and prediction performance. These results suggest that swarmDMD can be a powerful tool in analyzing biological swarms and engineering intelligent multi-agent systems. 

There are a number of future directions that are suggested by this work.  
First, it may be possible to learn nonlinear interaction laws using nonlinear generalizations of DMD, such as the sparse identification of nonlinear dynamics (SINDy)~\cite{brunton_discovering_2016} or tensor DMD formulations~\cite{klus2018tensor,harris2021time}. 
Moreover, sparsity promoting algorithms may be useful for inferring minimal network connectivity and causality~\cite{stepaniants2020inferring}; sparsity-promoting DMD has already been introduced to identify a minimal set of modes~\cite{Jovanovic2014pof}, but may be adapted for sparse dynamics matrices.  
For higher order systems, or systems with partially observed dynamics, incorporating time delays will also be important~\cite{hirsh2021structured, Brunton2021koopman}. 
Similarly, if data is collected in partial and overlapping domains, it may be possible to merge these data by aligning the phase between different DMD models~\cite{nair2020phase}.  
It may also be important to cluster the swarm into distinct regions, where different local rules are applied, for example using cluster reduced order modeling~\cite{Kaiser2014jfm,Kaiser2018jcp}. 
Finite-time Lyapunov exponents and Lagrangian coherent structures may also provide further insights into swarm dynamics and behavioral regimes~\cite{Haller2002pof,Shadden2005pd,shadden:07,Haller2015arfm,gunnarson2021learning}. 
It will also be interesting to expore the effectiveness of swarmDMD models for active feedback control of the swarm, for example by manipulating the behavior of a small subset of agents.  
Finally, it will be important to apply these methods to other canonical swarm models and also to real-world data to explore the strengths and weaknesses of these various algorithms.   

\section*{Acknowledgments}
SLB and ZS would like to thank the National Science Foundation (NSF) for partial funding under award AI Institute in Dynamic Systems (CBET-2112085). SLB would like to acknowledge the Air Force Office of Scientific Research (AFOSR FA9550-21-1-0178).  ZS would also like to thank NSF's support under award EPSCoR Research Infrastructure (OIA-2032522).

\section*{Code Availability}
The code for this work has been made available on GitHub at \url{https://github.com/e-vic/swarmDMD}.

%%%%%%%%%%%
%% BIBLIOGRAPHY
%%%%%%%%%%%
 \bibliographystyle{IEEEtran}
 \bibliography{references}

% Generated by IEEEtran.bst, version: 1.14 (2015/08/26)
\begin{thebibliography}{10}
\providecommand{\url}[1]{#1}
\csname url@samestyle\endcsname
\providecommand{\newblock}{\relax}
\providecommand{\bibinfo}[2]{#2}
\providecommand{\BIBentrySTDinterwordspacing}{\spaceskip=0pt\relax}
\providecommand{\BIBentryALTinterwordstretchfactor}{4}
\providecommand{\BIBentryALTinterwordspacing}{\spaceskip=\fontdimen2\font plus
\BIBentryALTinterwordstretchfactor\fontdimen3\font minus
  \fontdimen4\font\relax}
\providecommand{\BIBforeignlanguage}[2]{{%
\expandafter\ifx\csname l@#1\endcsname\relax
\typeout{** WARNING: IEEEtran.bst: No hyphenation pattern has been}%
\typeout{** loaded for the language `#1'. Using the pattern for}%
\typeout{** the default language instead.}%
\else
\language=\csname l@#1\endcsname
\fi
#2}}
\providecommand{\BIBdecl}{\relax}
\BIBdecl

\bibitem{Bonabeau:99a}
E.~Bonabeau, M.~Dorigo, and G.~Theraulaz, \emph{Swarm Intelligence: From
  Natural to Artificial Systems}, ser. Santa Fe Institute Studies on the
  Sciences of Complexity.\hskip 1em plus 0.5em minus 0.4em\relax Oxford
  University Press, 1999.

\bibitem{Lukeman2010}
R.~Lukeman, Y.-X. Li, and L.~Edelstein-Keshet, ``{Inferring individual rules
  from collective behavior},'' \emph{Proceedings of the National Academy of
  Sciences}, vol. 107, no.~28, pp. 12\,576--12\,580, jul 2010.

\bibitem{Bialek2014}
\BIBentryALTinterwordspacing
W.~Bialek, A.~Cavagna, I.~Giardina, T.~Mora, O.~Pohl, E.~Silvestri, M.~Viale,
  and A.~M. Walczak, ``Social interactions dominate speed control in poising
  natural flocks near criticality,'' \emph{Proceedings of the National Academy
  of Sciences}, vol. 111, no.~20, pp. 7212--7217, May 2014. [Online].
  Available: \url{https://doi.org/10.1073/pnas.1324045111}
\BIBentrySTDinterwordspacing

\bibitem{Berman2011}
S.~Berman, Q.~Lindsey, M.~S. Sakar, V.~Kumar, and S.~C. Pratt, ``{Experimental
  study and modeling of group retrieval in ants as an approach to collective
  transport in swarm robotic systems},'' \emph{Proceedings of the IEEE},
  vol.~99, no.~9, pp. 1470--1481, 2011.

\bibitem{Vicsek2012}
\BIBentryALTinterwordspacing
T.~Vicsek and A.~Zafeiris, ``{Collective motion},'' \emph{Physics Reports},
  vol. 517, no. 3-4, pp. 71--140, 2012. [Online]. Available:
  \url{http://dx.doi.org/10.1016/j.physrep.2012.03.004}
\BIBentrySTDinterwordspacing

\bibitem{Brambilla2013}
M.~Brambilla, E.~Ferrante, M.~Birattari, and M.~Dorigo, ``{Swarm robotics: A
  review from the swarm engineering perspective},'' \emph{Swarm Intelligence},
  vol.~7, no.~1, pp. 1--41, 2013.

\bibitem{Kelley2013}
D.~H. Kelley and N.~T. Ouellette, ``Emergent dynamics of laboratory insect
  swarms,'' \emph{Scientific Reports}, vol.~3, p. 1073, 2013.

\bibitem{Strombom2014}
\BIBentryALTinterwordspacing
D.~Str{\"{o}}mbom, R.~P. Mann, A.~M. Wilson, S.~Hailes, A.~J. Morton, D.~J.~T.
  Sumpter, and A.~J. King, ``{Solving the shepherding problem: heuristics for
  herding autonomous, interacting agents},'' \emph{Journal of The Royal Society
  Interface}, vol.~11, no. 100, p. 20140719, 2014. [Online]. Available:
  \url{https://royalsocietypublishing.org/doi/abs/10.1098/rsif.2014.0719}
\BIBentrySTDinterwordspacing

\bibitem{Paranjape2018}
A.~A. Paranjape, S.-J. Chung, K.~Kim, and D.~H. Shim, ``Robotic herding of a
  flock of birds using an unmanned aerial vehicle,'' \emph{IEEE Transactions on
  Robotics}, vol.~34, no.~4, pp. 901--915, 2018.

\bibitem{Romano2019}
\BIBentryALTinterwordspacing
D.~Romano, E.~Donati, G.~Benelli, and C.~Stefanini, ``{A review on
  animal–robot interaction: from bio-hybrid organisms to mixed societies},''
  \emph{Biological Cybernetics}, vol. 113, no.~3, pp. 201--225, 2019. [Online].
  Available: \url{https://doi.org/10.1007/s00422-018-0787-5}
\BIBentrySTDinterwordspacing

\bibitem{Prorok2011}
A.~Prorok, N.~Correll, and A.~Martinoli, ``{Multi-level spatial modeling for
  stochastic distributed robotic systems},'' \emph{The International Journal of
  Robotics Research}, vol.~30, no.~5, pp. 574--589, 2011.

\bibitem{Rubenstein2014}
M.~Rubenstein, A.~Cornejo, and R.~Nagpal, ``{Robotics. Programmable
  self-assembly in a thousand-robot swarm.}'' \emph{Science (New York, N.Y.)},
  vol. 345, no. 6198, pp. 795--9, 2014.

\bibitem{Chung2018}
\BIBentryALTinterwordspacing
S.-J. Chung, A.~A. Paranjape, P.~Dames, S.~Shen, and V.~Kumar, ``A survey on
  aerial swarm robotics,'' \emph{IEEE Transactions on Robotics}, vol.~34,
  no.~4, pp. 837--855, aug 2018. [Online]. Available:
  \url{https://ieeexplore.ieee.org/document/8424838/}
\BIBentrySTDinterwordspacing

\bibitem{McGuire2019}
\BIBentryALTinterwordspacing
K.~N. McGuire, C.~D. Wagter, K.~Tuyls, H.~J. Kappen, and G.~C. H.~E. de~Croon,
  ``{Minimal navigation solution for a swarm of tiny flying robots to explore
  an unknown environment},'' \emph{Science Robotics}, vol.~4, no.~35, p.
  eaaw9710, 2019. [Online]. Available:
  \url{https://www.science.org/doi/abs/10.1126/scirobotics.aaw9710}
\BIBentrySTDinterwordspacing

\bibitem{Schranz2020}
\BIBentryALTinterwordspacing
M.~Schranz, M.~Umlauft, M.~Sende, and W.~Elmenreich, ``Swarm robotic behaviors
  and current applications,'' \emph{Frontiers in Robotics and AI}, vol.~7,
  2020. [Online]. Available:
  \url{https://www.frontiersin.org/article/10.3389/frobt.2020.00036}
\BIBentrySTDinterwordspacing

\bibitem{Berlinger2021}
F.~Berlinger, M.~Gauci, and R.~Nagpal, ``{Implicit coordination for 3D
  underwater collective behaviors in a fish-inspired robot swarm},''
  \emph{Science Robotics}, vol.~6, no.~50, p. eabd8668, 2021.

\bibitem{Dorigo2021}
M.~Dorigo, G.~Theraulaz, and V.~Trianni, ``Swarm robotics: Past, present, and
  future [point of view],'' \emph{Proceedings of the IEEE}, vol. 109, no.~7,
  pp. 1152--1165, 2021.

\bibitem{reynolds_flocks_1987}
\BIBentryALTinterwordspacing
C.~W. Reynolds, ``\BIBforeignlanguage{en}{Flocks, herds and schools: {A}
  distributed behavioral model},'' \emph{\BIBforeignlanguage{en}{ACM SIGGRAPH
  Computer Graphics}}, vol.~21, no.~4, pp. 25--34, Aug. 1987. [Online].
  Available: \url{https://dl.acm.org/doi/10.1145/37402.37406}
\BIBentrySTDinterwordspacing

\bibitem{Spears2004}
W.~M. Spears, D.~F. Spears, J.~C. Hamann, and R.~Heil, ``{Distributed,
  physics-based control of swarms of vehicles},'' \emph{Autonomous Robots},
  vol.~17, no. 2-3, pp. 137--162, 2004.

\bibitem{Olfati-saber2006}
R.~Olfati-Saber, ``Flocking for multi-agent dynamic systems: Algorithms and
  theory,'' \emph{IEEE Transactions on Automatic Control}, vol.~51, no.~3, pp.
  401--420, mar 2006.

\bibitem{Hsieh2008}
M.~A. Hsieh, V.~Kumar, and L.~Chaimowicz, ``{Decentralized controllers for
  shape generation with robotic swarms},'' \emph{Robotica}, vol.~26, no.~26,
  pp. 691--701, 2008.

\bibitem{Pimenta2013}
L.~C.~A. Pimenta, G.~A.~S. Pereira, N.~Michael, R.~C. Mesquita, M.~M. Bosque,
  L.~Chaimowicz, and V.~Kumar, ``{Swarm coordination based on smoothed particle
  hydrodynamics technique},'' \emph{IEEE Transactions on Robotics}, vol.~29,
  no.~2, pp. 383--399, 2013.

\bibitem{Song2017}
Z.~Song, D.~Lipinski, and K.~Mohseni, ``{Multi-vehicle cooperation and nearly
  fuel-optimal flock guidance in strong background flows},'' \emph{Ocean
  Engineering}, vol. 141, pp. 388--404, sep 2017.

\bibitem{schmidt_distilling_2009}
M.~Schmidt and H.~Lipson, ``Distilling free-form natural laws from experimental
  data,'' \emph{Science}, vol. 324, no. 5923, pp. 81--85, 2009.

\bibitem{brunton_discovering_2016}
S.~L. Brunton, J.~L. Proctor, and J.~N. Kutz,
  ``\BIBforeignlanguage{en}{Discovering governing equations from data by sparse
  identification of nonlinear dynamical systems},''
  \emph{\BIBforeignlanguage{en}{Proceedings of the National Academy of
  Sciences}}, vol. 113, no.~15, pp. 3932--3937, Apr. 2016.

\bibitem{kutz_dynamic_2016}
J.~N. Kutz, S.~L. Brunton, B.~W. Brunton, and J.~L. Proctor, \emph{Dynamic Mode
  Decomposition: Data-driven Modeling of Complex Systems}.\hskip 1em plus 0.5em
  minus 0.4em\relax SIAM, 2016.

\bibitem{Huttenrauch:19}
\BIBentryALTinterwordspacing
M.~H{{\"u}}ttenrauch, A.~{\v{S}}o{\v{s}}i{{\'c}}, and G.~Neumann, ``Deep
  reinforcement learning for swarm systems,'' \emph{Journal of Machine Learning
  Research}, vol.~20, no.~54, pp. 1--31, 2019. [Online]. Available:
  \url{http://jmlr.org/papers/v20/18-476.html}
\BIBentrySTDinterwordspacing

\bibitem{Zhong2020}
\BIBentryALTinterwordspacing
M.~Zhong, J.~Miller, and M.~Maggioni, ``{Data-driven discovery of emergent
  behaviors in collective dynamics},'' \emph{Physica D: Nonlinear Phenomena},
  vol. 411, p. 132542, oct 2020. [Online]. Available:
  \url{https://doi.org/10.1016/j.physd.2020.132542
  https://linkinghub.elsevier.com/retrieve/pii/S0167278919308152}
\BIBentrySTDinterwordspacing

\bibitem{davidson_hierarchical_2021}
\BIBentryALTinterwordspacing
J.~D. Davidson, M.~Vishwakarma, and M.~L. Smith, ``Hierarchical approach for
  comparing collective behavior across scales: {Cellular} systems to honey bee
  colonies,'' \emph{Frontiers in Ecology and Evolution}, vol.~9, p. 581222,
  Feb. 2021. [Online]. Available:
  \url{https://www.frontiersin.org/articles/10.3389/fevo.2021.581222/full}
\BIBentrySTDinterwordspacing

\bibitem{ling_collective_2019}
\BIBentryALTinterwordspacing
H.~Ling, G.~E. Mclvor, J.~Westley, K.~van~der Vaart, J.~Yin, R.~T. Vaughan,
  A.~Thornton, and N.~T. Ouellette, ``\BIBforeignlanguage{en}{Collective turns
  in jackdaw flocks: kinematics and information transfer},''
  \emph{\BIBforeignlanguage{en}{Journal of the Royal Society Interface}},
  vol.~16, no. 159, p. 20190450, Oct. 2019. [Online]. Available:
  \url{https://royalsocietypublishing.org/doi/10.1098/rsif.2019.0450}
\BIBentrySTDinterwordspacing

\bibitem{fujii_physically-interpretable_2020}
\BIBentryALTinterwordspacing
K.~Fujii, N.~Takeishi, M.~Hojo, Y.~Inaba, and Y.~Kawahara,
  ``\BIBforeignlanguage{en}{Physically-interpretable classification of
  biological network dynamics for complex collective motions},''
  \emph{\BIBforeignlanguage{en}{Scientific Reports}}, vol.~10, no.~1, p. 3005,
  Dec. 2020. [Online]. Available:
  \url{http://www.nature.com/articles/s41598-020-58064-w}
\BIBentrySTDinterwordspacing

\bibitem{sridhar_geometry_2021}
\BIBentryALTinterwordspacing
V.~H. Sridhar, L.~Li, D.~Gorbonos, M.~Nagy, B.~R. Schell, T.~Sorochkin, N.~S.
  Gov, and I.~D. Couzin, ``\BIBforeignlanguage{en}{The geometry of
  decision-making in individuals and collectives},''
  \emph{\BIBforeignlanguage{en}{Proceedings of the National Academy of
  Sciences}}, vol. 118, no.~50, p. e2102157118, Dec. 2021. [Online]. Available:
  \url{http://www.pnas.org/lookup/doi/10.1073/pnas.2102157118}
\BIBentrySTDinterwordspacing

\bibitem{van_der_vaart_environmental_2020}
\BIBentryALTinterwordspacing
K.~van~der Vaart, M.~Sinhuber, A.~M. Reynolds, and N.~T. Ouellette,
  ``\BIBforeignlanguage{en}{Environmental perturbations induce correlations in
  midge swarms},'' \emph{\BIBforeignlanguage{en}{Journal of The Royal Society
  Interface}}, vol.~17, no. 164, p. 20200018, Mar. 2020. [Online]. Available:
  \url{https://royalsocietypublishing.org/doi/10.1098/rsif.2020.0018}
\BIBentrySTDinterwordspacing

\bibitem{tang_genetic_2020}
\BIBentryALTinterwordspacing
W.~Tang, J.~D. Davidson, G.~Zhang, K.~E. Conen, J.~Fang, F.~Serluca, J.~Li,
  X.~Xiong, M.~Coble, T.~Tsai, G.~Molind, C.~H. Fawcett, E.~Sanchez, P.~Zhu,
  I.~D. Couzin, and M.~C. Fishman, ``\BIBforeignlanguage{en}{Genetic control of
  collective behavior in zebrafish},''
  \emph{\BIBforeignlanguage{en}{iScience}}, vol.~23, no.~3, p. 100942, Mar.
  2020. [Online]. Available:
  \url{https://linkinghub.elsevier.com/retrieve/pii/S2589004220301267}
\BIBentrySTDinterwordspacing

\bibitem{Sumpter:10}
D.~J.~T. Sumpter, \emph{Collective Animal Behavior}.\hskip 1em plus 0.5em minus
  0.4em\relax Princeton University Press, 2010.

\bibitem{vicsek_novel_1995}
\BIBentryALTinterwordspacing
T.~Vicsek, A.~Czirók, E.~Ben-Jacob, I.~Cohen, and O.~Shochet,
  ``\BIBforeignlanguage{en}{Novel type of phase transition in a system of
  self-driven particles},'' \emph{\BIBforeignlanguage{en}{Physical Review
  Letters}}, vol.~75, no.~6, pp. 1226--1229, Aug. 1995. [Online]. Available:
  \url{https://link.aps.org/doi/10.1103/PhysRevLett.75.1226}
\BIBentrySTDinterwordspacing

\bibitem{costanzo_spontaneous_2018}
\BIBentryALTinterwordspacing
A.~Costanzo and C.~K. Hemelrijk, ``Spontaneous emergence of milling (vortex
  state) in a {Vicsek}-like model,'' \emph{Journal of Physics D: Applied
  Physics}, vol.~51, no.~13, p. 134004, Apr. 2018. [Online]. Available:
  \url{https://iopscience.iop.org/article/10.1088/1361-6463/aab0d4}
\BIBentrySTDinterwordspacing

\bibitem{aoki_simulation_1982}
I.~Aoki, ``\BIBforeignlanguage{en}{A simulation study on the schooling
  mechanism in fish},'' \emph{\BIBforeignlanguage{en}{Nippon Suisan
  Gakkaishi}}, vol.~48, no.~8, pp. 1081--1088, 1982.

\bibitem{helbing_social_1995}
\BIBentryALTinterwordspacing
D.~Helbing and P.~Molnár, ``Social force model for pedestrian dynamics,''
  \emph{Physical Review E}, vol.~51, no.~5, pp. 4282--4286, May 1995. [Online].
  Available: \url{https://link.aps.org/doi/10.1103/PhysRevE.51.4282}
\BIBentrySTDinterwordspacing

\bibitem{schaeffer_extracting_2018}
\BIBentryALTinterwordspacing
H.~Schaeffer, G.~Tran, and R.~Ward, ``\BIBforeignlanguage{en}{Extracting sparse
  high-dimensional dynamics from limited data},''
  \emph{\BIBforeignlanguage{en}{SIAM Journal on Applied Mathematics}}, vol.~78,
  no.~6, pp. 3279--3295, Jan. 2018. [Online]. Available:
  \url{https://epubs.siam.org/doi/10.1137/18M116798X}
\BIBentrySTDinterwordspacing

\bibitem{sinhuber_equation_2021}
\BIBentryALTinterwordspacing
M.~Sinhuber, K.~van~der Vaart, Y.~Feng, A.~M. Reynolds, and N.~T. Ouellette,
  ``\BIBforeignlanguage{en}{An equation of state for insect swarms},''
  \emph{\BIBforeignlanguage{en}{Scientific Reports}}, vol.~11, no.~1, p. 3773,
  Dec. 2021. [Online]. Available:
  \url{http://www.nature.com/articles/s41598-021-83303-z}
\BIBentrySTDinterwordspacing

\bibitem{bhaskar_analyzing_2019}
D.~Bhaskar, A.~Manhart, J.~Milzman, J.~T. Nardini, K.~M. Storey, C.~M. Topaz,
  and L.~Ziegelmeier, ``\BIBforeignlanguage{en}{Analyzing collective motion
  with machine learning and topology},'' \emph{\BIBforeignlanguage{en}{Chaos:
  An Interdisciplinary Journal of Nonlinear Science}}, vol.~29, no.~12, p.
  123125, Dec. 2019.

\bibitem{mei_dynamic_2018}
\BIBentryALTinterwordspacing
W.~Mei, N.~E. Friedkin, K.~Lewis, and F.~Bullo, ``Dynamic models of appraisal
  networks explaining collective learning,'' \emph{IEEE Transactions on
  Automatic Control}, vol.~63, no.~9, pp. 2898--2912, Sep. 2018. [Online].
  Available: \url{https://ieeexplore.ieee.org/document/8115275/}
\BIBentrySTDinterwordspacing

\bibitem{lu_nonparametric_2019}
\BIBentryALTinterwordspacing
F.~Lu, M.~Zhong, S.~Tang, and M.~Maggioni,
  ``\BIBforeignlanguage{en}{Nonparametric inference of interaction laws in
  systems of agents from trajectory data},''
  \emph{\BIBforeignlanguage{en}{Proceedings of the National Academy of
  Sciences}}, vol. 116, no.~29, pp. 14\,424--14\,433, Jul. 2019. [Online].
  Available: \url{http://www.pnas.org/lookup/doi/10.1073/pnas.1822012116}
\BIBentrySTDinterwordspacing

\bibitem{schmid_dynamic_2010}
P.~J. Schmid, ``Dynamic mode decomposition of numerical and experimental
  data,'' \emph{Journal of Fluid Mechanics}, vol. 656, pp. 5--28, 2010.

\bibitem{Rowley2009jfm}
C.~W. Rowley, I.~Mezi\'c, S.~Bagheri, P.~Schlatter, and D.~Henningson,
  ``Spectral analysis of nonlinear flows,'' \emph{Journal Fluid Mechanics},
  vol. 645, pp. 115--127, 2009.

\bibitem{schmid2011applications}
P.~Schmid, L.~Li, M.~Juniper, and O.~Pust, ``Applications of the dynamic mode
  decomposition,'' \emph{Theoretical and Computational Fluid Dynamics},
  vol.~25, no. 1-4, pp. 249--259, 2011.

\bibitem{schmid2022dynamic}
P.~J. Schmid, ``Dynamic mode decomposition and its variants,'' \emph{Annual
  Review of Fluid Mechanics}, vol.~54, pp. 225--254, 2022.

\bibitem{Brunton2021koopman}
S.~L. Brunton, M.~Budi{\v{s}}i{\'c}, E.~Kaiser, and J.~N. Kutz, ``Modern
  {K}oopman theory for dynamical systems,'' \emph{arXiv preprint
  arXiv:2102.12086}, 2021.

\bibitem{proctor_dynamic_2016}
J.~L. Proctor, S.~L. Brunton, and J.~N. Kutz, ``\BIBforeignlanguage{en}{Dynamic
  mode decomposition with control},'' \emph{\BIBforeignlanguage{en}{SIAM
  Journal on Applied Dynamical Systems}}, vol.~15, no.~1, pp. 142--161, Jan.
  2016.

\bibitem{brunton2016extracting}
B.~W. Brunton, L.~A. Johnson, J.~G. Ojemann, and J.~N. Kutz, ``Extracting
  spatial--temporal coherent patterns in large-scale neural recordings using
  dynamic mode decomposition,'' \emph{Journal of Neuroscience Methods}, vol.
  258, pp. 1--15, 2016.

\bibitem{kunert2019extracting}
J.~M. Kunert-Graf, K.~M. Eschenburg, D.~J. Galas, J.~N. Kutz, S.~D. Rane, and
  B.~W. Brunton, ``Extracting reproducible time-resolved resting state networks
  using dynamic mode decomposition,'' \emph{Frontiers in Computational
  Neuroscience}, p.~75, 2019.

\bibitem{hirsh2021structured}
S.~M. Hirsh, S.~M. Ichinaga, S.~L. Brunton, J.~Nathan~Kutz, and B.~W. Brunton,
  ``Structured time-delay models for dynamical systems with connections to
  frenet--serret frame,'' \emph{Proceedings of the Royal Society A}, vol. 477,
  no. 2254, p. 20210097, 2021.

\bibitem{Brunton2019book}
S.~L. Brunton and J.~N. Kutz, \emph{Data-Driven Science and Engineering:
  Machine Learning, Dynamical Systems, and Control}.\hskip 1em plus 0.5em minus
  0.4em\relax Cambridge University Press, 2019.

\bibitem{Taira2017aiaa}
K.~Taira, S.~L. Brunton, S.~Dawson, C.~W. Rowley, T.~Colonius, B.~J. McKeon,
  O.~T. Schmidt, S.~Gordeyev, V.~Theofilis, and L.~S. Ukeiley, ``Modal analysis
  of fluid flows: An overview,'' \emph{AIAA Journal}, vol.~55, no.~12, pp.
  4013--4041, 2017.

\bibitem{Taira2020aiaaj}
K.~Taira, M.~S. Hemati, S.~L. Brunton, Y.~Sun, K.~Duraisamy, S.~Bagheri,
  S.~Dawson, and C.-A. Yeh, ``Modal analysis of fluid flows: Applications and
  outlook,'' \emph{AIAA Journal}, vol.~58, no.~3, pp. 998--1022, 2020.

\bibitem{Katz2011}
\BIBentryALTinterwordspacing
Y.~Katz, K.~Tunstrom, C.~C. Ioannou, C.~Huepe, and I.~D. Couzin, ``Inferring
  the structure and dynamics of interactions in schooling fish,''
  \emph{Proceedings of the National Academy of Sciences}, vol. 108, no.~46, pp.
  18\,720--18\,725, Jul. 2011. [Online]. Available:
  \url{https://doi.org/10.1073/pnas.1107583108}
\BIBentrySTDinterwordspacing

\bibitem{klus2018tensor}
S.~Klus, P.~Gel{\ss}, S.~Peitz, and C.~Sch{\"u}tte, ``Tensor-based dynamic mode
  decomposition,'' \emph{Nonlinearity}, vol.~31, no.~7, p. 3359, 2018.

\bibitem{harris2021time}
K.~D. Harris, A.~Aravkin, R.~Rao, and B.~W. Brunton, ``Time-varying
  autoregression with low-rank tensors,'' \emph{SIAM Journal on Applied
  Dynamical Systems}, vol.~20, no.~4, pp. 2335--2358, 2021.

\bibitem{stepaniants2020inferring}
G.~Stepaniants, B.~W. Brunton, and J.~N. Kutz, ``Inferring causal networks of
  dynamical systems through transient dynamics and perturbation,''
  \emph{Physical Review E}, vol. 102, no.~4, p. 042309, 2020.

\bibitem{Jovanovic2014pof}
M.~R. Jovanovi{\'c}, P.~J. Schmid, and J.~W. Nichols, ``Sparsity-promoting
  dynamic mode decomposition,'' \emph{Physics of Fluids}, vol.~26, no.~2, p.
  024103, 2014.

\bibitem{nair2020phase}
A.~G. Nair, B.~Strom, B.~W. Brunton, and S.~L. Brunton, ``Phase-consistent
  dynamic mode decomposition from multiple overlapping spatial domains,''
  \emph{Physical Review Fluids}, vol.~5, no.~7, p. 074702, 2020.

\bibitem{Kaiser2014jfm}
E.~Kaiser, B.~R. Noack, L.~Cordier, A.~Spohn, M.~Segond, M.~Abel, G.~Daviller,
  J.~Osth, S.~Krajnovic, and R.~K. Niven, ``Cluster-based reduced-order
  modelling of a mixing layer,'' \emph{Journal of Fluid Mechanics}, vol. 754,
  pp. 365--414, 2014.

\bibitem{Kaiser2018jcp}
E.~Kaiser, M.~Morzynski, G.~Daviller, J.~N. Kutz, B.~W. Brunton, and S.~L.
  Brunton, ``Sparsity enabled cluster reduced-order models for control,''
  \emph{Journal of Computational Physics}, vol. 352, pp. 388--409, 2018.

\bibitem{Haller2002pof}
G.~Haller, ``{Lagrangian} coherent structures from approximate velocity data,''
  \emph{Physics of Fluids}, vol.~14, no.~6, pp. 1851--1861, June 2002.

\bibitem{Shadden2005pd}
S.~C. Shadden, F.~Lekien, and J.~E. Marsden, ``Definition and properties of
  {Lagrangian} coherent structures from finite-time {Lyapunov} exponents in
  two-dimensional aperiodic flows,'' \emph{Physica D}, vol. 212, pp. 271--304,
  2005.

\bibitem{shadden:07}
S.~C. Shadden, K.~Katija, M.~Rosenfeld, J.~E. Marsden, and J.~O. Dabiri,
  ``Transport and stirring induced by vortex formation,'' \emph{Journal of
  Fluid Mechanics}, vol. 593, pp. 315--331, 2007.

\bibitem{Haller2015arfm}
G.~Haller, ``Lagrangian coherent structures,'' \emph{Annual Review of Fluid
  Mechanics}, vol.~47, pp. 137--162, 2015.

\bibitem{gunnarson2021learning}
P.~Gunnarson, I.~Mandralis, G.~Novati, P.~Koumoutsakos, and J.~O. Dabiri,
  ``Learning efficient navigation in vortical flow fields,'' \emph{arXiv
  preprint arXiv:2102.10536}, 2021.

\end{thebibliography}

\appendix

Here we will briefly discuss the performance of standard swarmDMD when trained on the ground truth cases where $\eta=\pi/12$, and of the first-order (FO) Cartesian and polar implementations. 

\begin{figure*}
\vspace{-.2in}
     \centering
     \includegraphics[width=\textwidth]{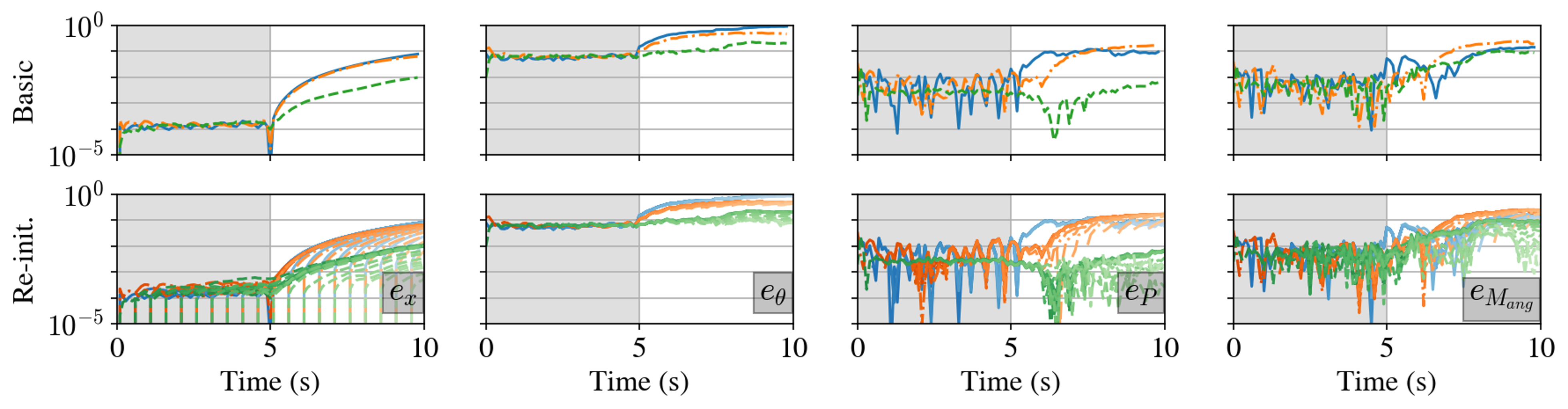}
     
    \caption{From left to right, the resulting position, heading, polarisation, and angular momentum errors of the standard dynamics are compared across ground truth interaction radii, $r$, 0.05 (blue), 0.25 (orange), and 0.5 (green). Ground truth has $\eta=\pi/12$ for basic and re-initialisation.}
    \label{fig:errors_noisy}
\end{figure*}

\begin{figure*}[t]
\vspace{-.1in}
    \centering    
    \begin{minipage}{0.48\linewidth}
    \begin{centering}
        \includegraphics[width=\linewidth]{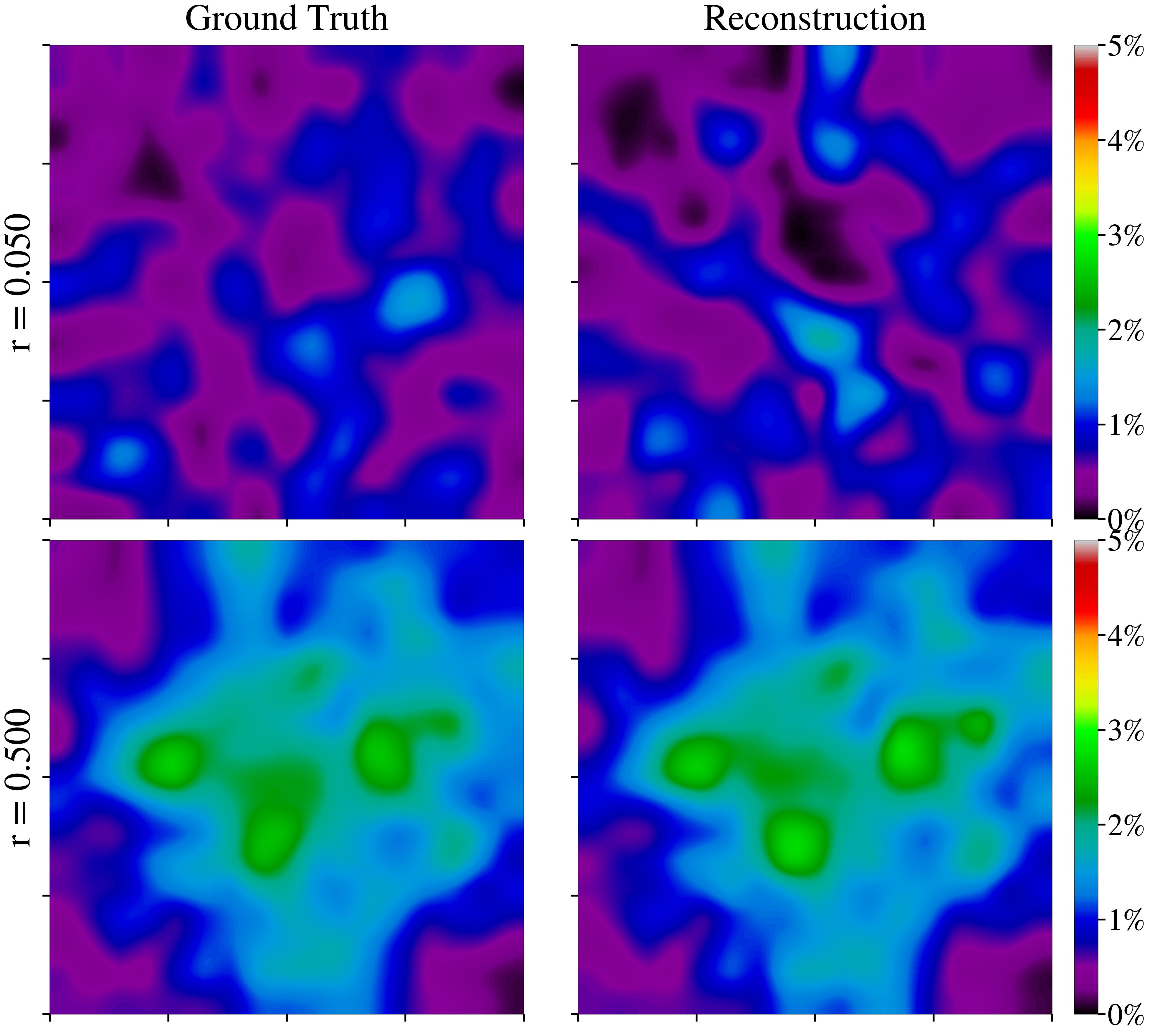}
        \caption{Agent distribution analysis of swarmDMD with standard dynamics over the training period, ground truth has $\eta=\pi/12$. On the left is the ground truth distribution, and on the right is the swarmDMD recreation. }
        \label{fig:standard_densityT_noisy}
    \end{centering}
    \end{minipage}\hspace{0.3cm}%
    \begin{minipage}{0.48\linewidth}
    \begin{centering}
        \includegraphics[width=\linewidth]{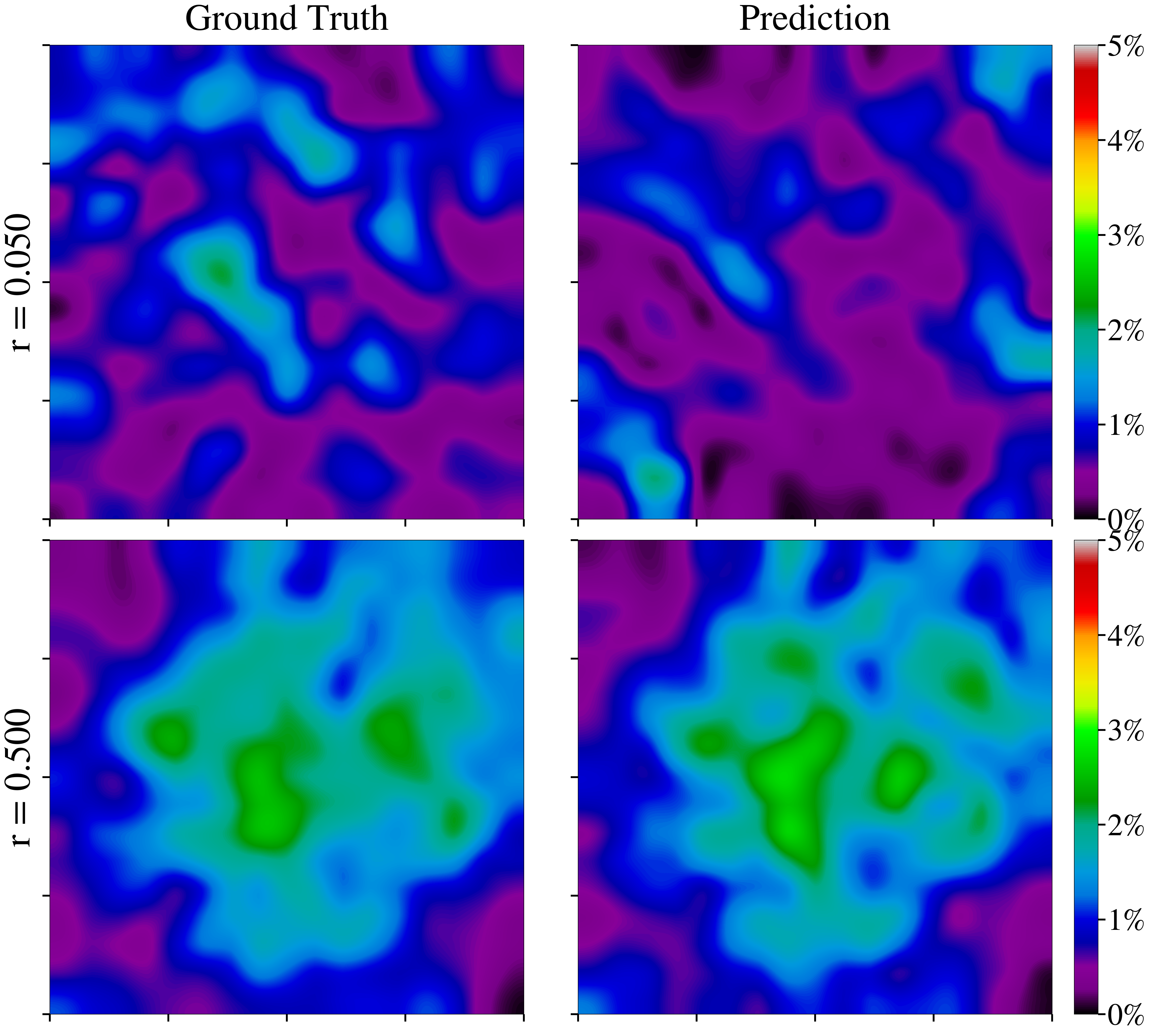}
        \caption{Agent distribution analysis of swarmDMD with standard dynamics over a period of 5s prediction, ground truth has $\eta=\pi/12$. On the left is the ground truth distribution, and on the right is the swarmDMD prediction. }
        \label{fig:standard_densityP_noisy}
    \end{centering}
    \end{minipage}
    \vspace{-.1in}
\end{figure*}

\subsection*{Standard Dynamics}
Position, heading, polarisation, and angular momentum --- seen in Figure \ref{fig:errors_noisy} --- all show increases in error values from the $\eta=0$ cases, with heading error increasing multiple orders of magnitude. Polarisation and angular momentum errors show a significant increase in noisiness of the plots. These are all expected results as the uncertainty in the headings of the ground truth agents is now non-zero. It is also interesting to note that with the increase in uncertainty in the ground truth swarm, changes in interaction radius seem to have less effect on the results.

% \begin{figure}
% \vspace{-.2in}
%      \centering
%      \begin{subfigure}{0.43\textwidth}
%          \centering
%          \includegraphics[width=\textwidth]{template/Figures/Position_eta02617993877991494.png}
%          \caption{Position error.}
%          \label{fig:position_noisy}
%      \end{subfigure}

%      \begin{subfigure}{0.43\textwidth}
%          \centering
%          \includegraphics[width=\textwidth]{template/Figures/Heading_eta02617993877991494.png}
%          \caption{Heading error.}
%          \label{fig:heading_noisy}
%      \end{subfigure}
     
%      \begin{subfigure}{0.43\textwidth}
%          \centering
%          \includegraphics[width=\textwidth]{template/Figures/Polarisation_eta02617993877991494.png}
%          \caption{Polarisation error.}
%          \label{fig:polarisation_noisy}
%      \end{subfigure}
     
%      \begin{subfigure}{0.43\textwidth}
%          \centering
%          \includegraphics[width=\textwidth]{template/Figures/AngularMomentum_eta02617993877991494.png}
%          \caption{Angular momentum error.}
%          \label{fig:momentum_noisy}
%      \end{subfigure}
     
%         \caption{The resulting position, heading, polarisation, and angular momentum errors of the standard dynamics are compared across ground truth interaction radii, $r$, 0.05 (blue), 0.25 (orange), and 0.5 (green). Ground truth has $\eta=\pi/12$ for basic and re-initialisation.}
%         \label{fig:errors_noisy}
%         \vspace{-.2in}
% \end{figure}

Figures \ref{fig:standard_densityT_noisy} and \ref{fig:standard_densityP_noisy} show the agent density distribution during the training and prediction periods, respectively. Even with increased uncertainty in the ground truth swarms, swarmDMD still does a reasonable job of capturing the structure of the agents within the swarm, with the general patterns showing up in both training and predictions for $r=0.05$, and the almost identical pattern during training for $r=0.5$. 

\subsection*{First-Order Dynamics}
\begin{figure*}[t]
\vspace{-.2in}
     \centering
     \includegraphics[width=\textwidth]{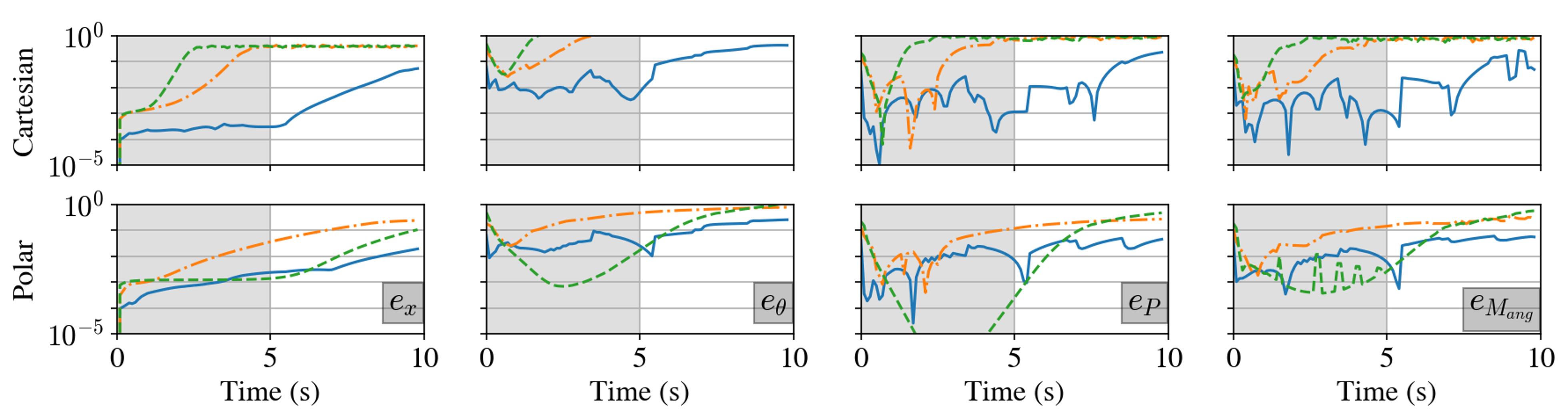}
    \caption{From left to right, the resulting position, heading, polarisation, and angular momentum errors of the first order Cartesian and polar dynamics compared across ground truth interaction radii, $r$, 0.05 (blue), 0.25 (orange), and 0.5 (green). Ground truth has $\eta=0$.}
    \label{fig:errors_FOs}
    \vspace{-.2in}
\end{figure*}

Figure \ref{fig:errors_FOs} shows the position, heading, polarisation, and angular momentum plots for the FO Cartesian and polar dynamics. 
In general, FO polar performs better than FO Cartesian. This is reasonable since FO Cartesian uses position and velocity information, which in the datatype analysis resulted in higher error than when relative distance or heading were used. 
%The time step size I also think is a limiting factor in the accuracy of the results, as including first order terms in the dynamics requires ``more integration", and without a sufficiently small time step this could result in large integration error
It is interesting to note that across error metrics the difference in error magnitude between training and prediction periods is not as significant as when the standard dynamics are used; the error often begins to increase during the training period.

\begin{figure}[b]
\vspace{-.2in}
    \centering    
    \begin{centering}
        \includegraphics[width=\linewidth]{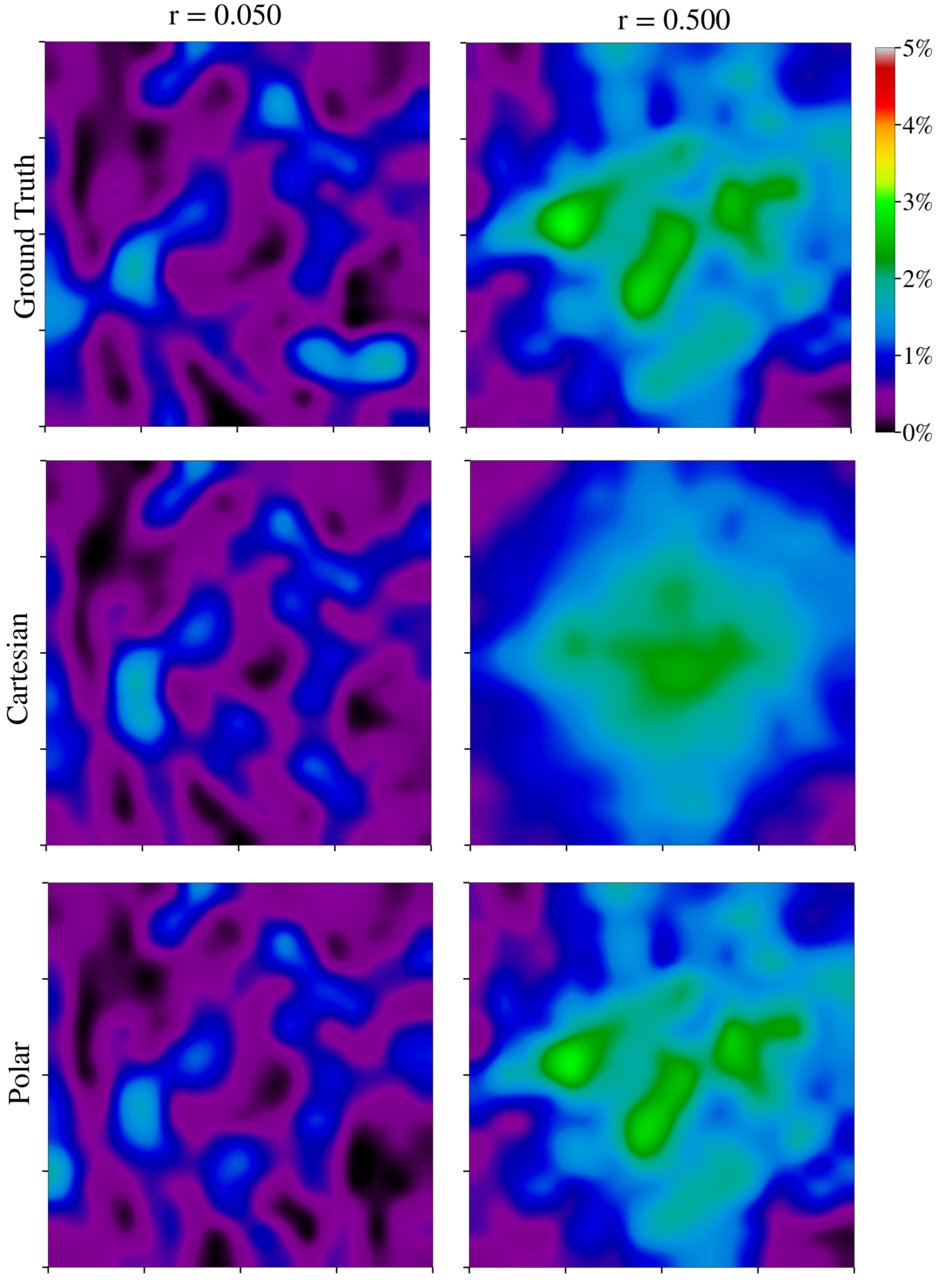}
        \caption{Agent distribution analysis of swarmDMD with FO Cartesian and polar dynamics over the training period, ground truth has $\eta=0$. On the left are the distributions for $r=0.05$, and on the right the distributions for $r=0.5$. The ground truth distributions are shown in the first row. }
        \label{fig:densityT_FO}
    \end{centering}
    \vspace{-.2in}
\end{figure}

\begin{figure}[b]
\vspace{-.2in}
    \centering    
    \begin{centering}
        \includegraphics[width=\linewidth]{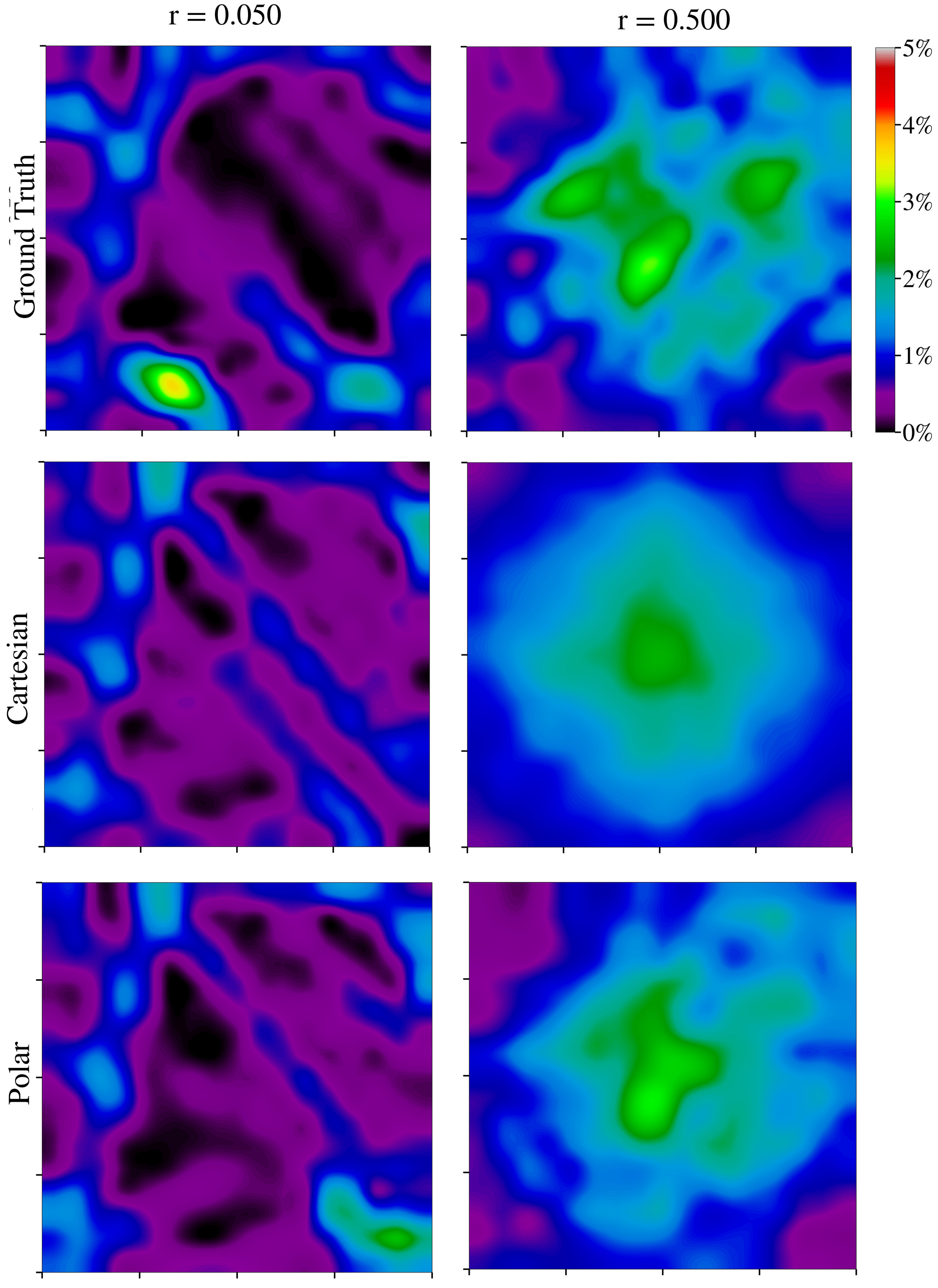}
        \caption{Agent distribution analysis of swarmDMD with FO Cartesian and polar dynamics over a period of 5s prediction, ground truth has $\eta=0$. On the left are the distributions for $r=0.05$, and on the right the distributions for $r=0.5$. The ground truth distributions are shown in the first row. }
        \label{fig:densityP_FO}
    \end{centering}
    \vspace{-.2in}
\end{figure}

Figures \ref{fig:densityT_FO} and \ref{fig:densityP_FO} give the agent density distribution for the FO Cartesian and polar dynamics. The FO polar distributions are almost identical for both the training and prediction periods, whereas the distributions for the FO Cartesian dynamics only capture the general pattern for training and prediction with $r=0.05$, and capture almost none of the pattern for $r=0.5$.

% \begin{figure*}[b]
%      \centering
%      \begin{subfigure}[b]{0.48\textwidth}
%          \centering
%          \includegraphics[width=\textwidth]{template/Figures/FO_Position_eta0.png}
%          \caption{Position error.}
%          \label{fig:position_altdynamics}
%      \end{subfigure}
%      \hfill
%      \begin{subfigure}[b]{0.48\textwidth}
%          \centering
%          \includegraphics[width=\textwidth]{template/Figures/FO_Heading_eta0.png}
%          \caption{Heading error.}
%          \label{fig:heading_altdynamics}
%      \end{subfigure}
     
%      \begin{subfigure}[b]{0.48\textwidth}
%          \centering
%          \includegraphics[width=\textwidth]{template/Figures/FO_Polarisation_eta0.png}
%          \caption{Polarisation error.}
%          \label{fig:polarisation_altdynamics}
%      \end{subfigure}
%      \hfill
%      \begin{subfigure}[b]{0.48\textwidth}
%          \centering
%          \includegraphics[width=\textwidth]{template/Figures/FO_AngularMomentum_eta0.png}
%          \caption{Angular momentum error.}
%          \label{fig:momentum_altdynamics}
%      \end{subfigure}
     
%         \caption{The resulting position, heading, polarisation, and angular momentum errors of the first order Cartesian and polar dynamics compared across ground truth interaction radii, $r$, 0.05 (blue), 0.25 (orange), and 0.5 (green). Ground truth has $\eta=0$.}
%         \label{fig:errors_FOs}
% \end{figure*}

\newpage

% \begin{figure*}[t]
% \vspace{-.2in}
%     \centering    
%     \begin{centering}
%         \includegraphics[width=0.8\linewidth]{template/Figures/NEW-DensityTraining_eta0_methodFOs.png}
%         \caption{Agent distribution analysis of swarmDMD with FO Cartesian and polar dynamics over the training period, ground truth has $\eta=0$. On the left is the ground truth distribution, centred is the FO Cartesian recreation, and on the right is the FO polar. }
%         \label{fig:densityT_FO}
%     \end{centering}
%     \vspace{-.2in}
% \end{figure*}

% \begin{figure*}[b]
% \vspace{-.2in}
%     \centering    
%     \begin{centering}
%         \includegraphics[width=0.8\linewidth]{template/Figures/NEW-DensityPrediction_eta0_methodFOs.png}
%         \caption{Agent distribution analysis of swarmDMD with FO Cartesian and polar dynamics over a period of 5s prediction, ground truth has $\eta=0$. On the left is the ground truth distribution, and the centre and right are the swarmDMD predictions using FO Cartesian and polar dynamics, respectively. }
%         \label{fig:densityP_FO}
%     \end{centering}
%     \vspace{-.2in}
% \end{figure*}

\vfill

\vfill

\end{document}